\documentclass[runningheads]{llncs}
\usepackage[T1]{fontenc}
\usepackage{graphicx}
\usepackage{subfigure}
\usepackage{mathtools}
\usepackage{algorithm}
\usepackage{algorithmicx}
\usepackage{algpseudocode}
\usepackage{threeparttable}
\usepackage{makecell}
\usepackage{amsmath,amssymb,amsfonts}

\newtheorem{Def}{Definition}
\newtheorem{The}{Theorem}
\newtheorem{Rem}{Remark}
\newtheorem{Lem}{Lemma}

\begin{document}
\title{Data Heterogeneity Differential Privacy: From Theory to Algorithm
	\thanks{This work was supported in part by the Excellent Talents Program of Institute of Information Engineering, CAS, the Special Research Assistant Project of CAS, the Beijing Outstanding Young Scientist Program (No. BJJWZYJH012019100020098), Beijing Natural Science Foundation (No. 4222029), and National Natural Science Foundation of China (No. 62076234, No. 62106257).}
}
%
%
\author{Yilin Kang\inst{1,3} \and
Jian Li\inst{1} \and
Yong Liu\inst{2} \and
Weiping Wang\inst{1}}
\authorrunning{Kang et al.}
%
\institute{Institute of Information Engineering, Chinese Academy of Sciences
\email{\{kangyilin,lijian9026,wangweiping\}@iie.ac.cn} \\
\and
Gaoling School of Artificial Intellignece, Renmin University of China
\email{liuyonggsai@ruc.edu.cn} \\
\and
School of Cyber Security, University of Chinese Academy of Sciences\\
}
\maketitle              
\begin{abstract}
Traditionally, the random noise is equally injected when training with different data instances in the field of differential privacy (DP).
In this paper, we first give sharper excess risk bounds of DP stochastic gradient descent (SGD) method.
Considering most of the previous methods are under convex conditions, we use Polyak-{\L}ojasiewicz condition to relax it in this paper.
Then, after observing that different training data instances affect the machine learning model to different extent, we consider the heterogeneity of training data and attempt to improve the performance of DP-SGD from a new perspective.
Specifically, by introducing the influence function (IF), we quantitatively measure the contributions of various training data on the final machine learning model.
If the contribution made by a single data instance is so little that attackers cannot infer anything from the model, we do not add noise when training with it.
Based on this observation, we design a `Performance Improving' DP-SGD algorithm: PIDP-SGD.
Theoretical and experimental results show that our proposed PIDP-SGD improves the performance significantly.

\end{abstract}
\section{Introduction}
Machine learning has been widely applied to many fields in recent decades and tremendous data has been collected.
As a result, information disclosure becomes a huge problem.
Except for the original data, model parameters can reveal sensitive information in an undirect way as well \cite{fredrikson2014privacy,shokri2017membership}.

Differential privacy (DP) \cite{dwork2006calibrating,dwork2014algorithmic} is a theoretically rigorous tool to prevent sensitive information \cite{chen2019differentially}.
It preserves privacy by introducing random noise, to block adversaries from inferring any single individual included in the dataset by observing the machine learning model.
As such, DP has attracted many researchers and has been applied to numerous machine learning methods \cite{shokri2015privacy,zhao2018inprivate,bernstein2019differentially,wang2019principal,heikkila2019differentially,xu2019ganobfuscator,ullman2019efficiently,arora2019differentially}.
Some kind of attacks, such as membership inference attack, attribute inference attack, memorization attack, can be thwarted by differential privacy \cite{backes2016membership,carlini2019secret,bargav2019evaluating}.
Empirical risk minimization (ERM), a popular optimization method that covers a variety of machine learning tasks, also faces privacy problems.
A long list of works focuses on applying differential privacy to ERM \cite{chaudhuri2011differentially,kifer2012private,song2013stochastic,bassily2014private,wang2017differentially,zhang2017efficient,phan2017adaptive,wu2017bolt,wang2019dp}, in which three main approaches are studied: output perturbation, objective perturbation, and gradient perturbation.
Although there are wildspread researches these years, some problems still exist:
Firstly, all data is usually treated equally when training DP model.
However, in real scenarios, different training data affects the model differently, so treating them all the same lacks `common sense' and is one of the reasons why low accuracy appears.
Meanwhile, previous results always require that the loss function is convex (or even strongly convex), so the application scenario is narrow.
Besides, existed excess risk bounds for non-convex setting is unsatisfactory.

To solve the problems, we make the following contributions in this paper:
First, we introduce the Polyak-{\L}ojasiewicz condition \cite{karimi2016linear} to relax the convex (strongly convex) assumption.
We analyze the excess empirical risk, the generalization error, the excess population risk and give corresponding bounds under this assumption.
Theoretical results show that our given excess risk bounds are better than previous non-convex ones and even better than some of the convex ones.
Second, motivated by the definition of DP, we provide a new perspective to improve the performance: treating different data instances differently.
In particular, we introduce the Influence Function (IF) \cite{koh2017understanding} to measure the contributions made by different data instances.
If the data instance $z$ contributes so little to the machine learning model that the attacker cannot infer anything (represented by the privacy budget $\epsilon$ in DP), we do not add noise when training with $z$, rather than treating all of the data instances as being the same.
In this way, we propose a `Performance Improving' algorithm: PIDP-SGD method to improve the model performance, by taking data heterogeneity into account.

The rest of the paper is organized as follows.
First, we introduce some related work in Section 2.
Preliminaries are presented in Section 3.
We analyze the excess risk bounds of traditional DP-SGD method and give sharper theoretical bounds in Section 4.
The `Performance Improving' DP-SGD algorithm is given and analyzed in Section 5.
In Section 6, we compare our proposed method with previous methods in detail.
The experimental results are shown in 7.
Finally, we conclude the paper in Section 8.

\section{Related Work}

The first method on DP-ERM is proposed in \cite{chaudhuri2011differentially}, in which output and objective perturbation methods are introduced.
Gradient perturbation is proposed in \cite{song2013stochastic}, in which DP-SGD is analyzed for the first time.
The accuracy of the objective perturbation method is improved by \cite{kifer2012private}.
The excess empirical risk bounds of the methods proposed in \cite{chaudhuri2011differentially} and \cite{kifer2012private} are improved by \cite{bassily2014private}.
An output perturbation method is introduced to DP-SGD in \cite{wu2017bolt}, in which a novel $\ell_2$ sensitivity of SGD is analyzed and better accuracy is achieved.
\cite{wang2017differentially} introduces Prox-SVRG \cite{xiao2014proximal} to DP and proposes DP-SVRG, in which optimal or near-optimal utility bounds are achieved.

Meanwhile, there are also some works concentrated on non-convex analysis.
DP is introduced to deep learning by \cite{abadi2016deep}, via gradient perturbation method, however, it focuses on the privacy but lacks utility analysis.
An output perturbation method is proposed in \cite{zhang2017efficient} under non-convex condition.
The Polyak-{\L}ojasiewicz condition is introduced in \cite{wang2017differentially} and the excess empirical risk of gradient perturbation method under non-convex condition is analyzed, however, the excess population risk is not discussed.
Aiming to achieve better performance, in \cite{phan2017adaptive}, more noise is added to those features less `relevant' to the final model.
A Laplace smooth operator is introduced to DP-SGD and a new method: DP-LSSGD is proposed in \cite{wang2019dp}, focusing on non-convex analysis.
The excess empirical risk bound and the excess population risk bound of DP model under non-convex condition are analyzed by \cite{wang2019differentially}, via Gradient Langevin Dynamics.
For non-convex condition, the theoretical results are always unsatisfactory.

All the works mentioned above are based on three traditional perturbation methods, all data instances are treated equally, which lacks `common sense' and leads unsatisfactory utility.
To solve the problems under both convex and non-convex conditions, we take data heterogeneity into account and propose a `Performance Improving' algorithm.
In this way, our method improves the performance of the DP model, which is superior to previous methods in the excess risk bounds.

\section{Preliminaries}

\subsection{Notations and Assumptions}

The loss function is defined as $\ell:\mathcal{C}\times\mathcal{D}\rightarrow\mathbb{R}$, where $\mathcal{C}$ is the parameter space and $\mathcal{D}$ is the data universe.
We assume that the parameter space is bounded, whose radius is $r$.
Supposing there are	$n$ data instances in the dataset $D=\{z_1,\cdots,z_n\}\in\mathcal{D}^n$, where $z_i$ are drawn i.i.d from the underlying distribution $\mathcal{P}$.
Besides, for each $z=(x,y)$, $x$ is the feature and $y$ is the label.
We assume $\|x\|_2\leq1$, i.e. $\mathcal{X}$ is the unit ball.
Moreover, for a vector $x=[x_1,\cdots,x_d]$, its $\ell_2$ norm is defined as: $\|x\|_2=\big(\sum_{i=1}^{d}x_i^2\big)^{1/2}$, and the $i^{th}$ element $x_i$ is represented by $[x]_i$.
In the following, we use $\nabla_{\cdot}$ to denote the gradient over $\cdot$.

In ERM, our goal is to find the optimal model that minimizes the empirical risk $L(\theta;D)=\frac{1}{n}\sum_{i=1}^{n}\ell\big(\theta,z_i\big)$ on dataset $D$, defined as: $\theta^*=\arg\min\big[L\big(\theta;D\big)\big]$, and $L(\theta^*;D)$ is represented by $L^*$.
Additionally, the population risk is defined as $L_\mathcal{P}\big(\theta\big)=\mathbb{E}_{z\sim\mathcal{P}}\big[\ell\big(\theta,z\big)\big]$.
For an algorithm $\mathcal{A}:\mathcal{D}^n\rightarrow\mathbb{R}^m$, we denote its output as $\theta_\mathcal{A}$.
The \textbf{excess empirical risk} denotes the gap between $\theta_\mathcal{A}$ and $\theta^*$, defined as: $L(\theta_\mathcal{A};D)-L^*$; and the \textbf{excess population risk} represents the gap between $\theta_\mathcal{A}$ and the optimal model over the underlying $\mathcal{P}$, defined as: $L_\mathcal{P}(\theta_\mathcal{A})-\min_{\theta\in\mathcal{C}}L_\mathcal{P}(\theta)$.
The \textbf{generalization error} connects the population risk and the empirical risk, defined as: $L_\mathcal{P}(\theta_\mathcal{A})-L(\theta_\mathcal{A};D)$.

Besides, there are some assumptions on the loss function:
\begin{Def}[$G$-Lipschitz]
	A loss function $\ell:\mathcal{C}\times\mathcal{D}\rightarrow\mathbb{R}$ is $G$-Lipschitz over $\theta$, if for some constant $G$, any $z\in\mathcal{D}$ and $\theta,\theta'\in\mathcal{C}$, we have: $|\ell(\theta,z)-\ell(\theta',z)|\leq G\|\theta-\theta'\|_2$.
\end{Def}

\begin{Def}[$L$-smooth]
	A loss function $\ell:\mathcal{C}\times\mathcal{D}\rightarrow\mathbb{R}$ is $L$-smooth over $\theta$, if for some constant $L$, any $z\in\mathcal{D}$ and $\theta,\theta'\in\mathcal{C}$, we have: $\|\nabla_\theta\ell(\theta,z)-\nabla_\theta\ell(\theta',z)\|_2\leq L\|\theta-\theta'\|_2$.
\end{Def}

\begin{Def}[$C$-Hessian Lipschitz]
	A loss function $\ell:\mathcal{C}\times\mathcal{D}\rightarrow\mathbb{R}$ is $C$-Hessian Lipschitz over $\theta$, if for any $z\in\mathcal{D}$ and $\theta,\theta'\in\mathcal{C}$, we have: $\|\nabla^2_\theta\ell(\theta,z)-\nabla^2_\theta\ell(\theta',z)\|_2\leq C\|\theta-\theta'\|_2$.
\end{Def}

\begin{Rem}
	$G$-Lipschitz means that $\|\nabla_{\theta}\ell(\theta,z)\|_2\leq G$.
	For Mean Squared Error, elements in $\nabla_\theta\ell(\theta,z)$ are less than $2c$ if $y-f(x)$\footnote{$f(x)$ is a general notation, for example, $f(x)=\theta^Tx$ in linear regression.} is bounded by $c$.
	For logistic regression over cross entropy, elements in $\nabla_\theta\ell(\theta,z)$ are less than 1.
	$L$-smooth means that $\|\nabla_{\theta}^2\ell(\theta,z)\|_2\leq L$.
	For Mean Squared Error, elements in $\nabla^2_\theta\ell(\theta,z)$ are less than 2.
	For logistic regression over cross entropy, elements in $\nabla^2_\theta\ell(\theta,z)$ are less than 0.25.
	$C$-Hessian Lipschitz means that $\|\nabla_{\theta}^3\ell(\theta,z)\|_2\leq C$.
	For Mean Squared Error, $\|\nabla^3_\theta\ell(\theta,z)\|_2=0$.
	For logistic regression over cross entropy, elements in $\nabla^3_\theta\ell(\theta,z)$ are less than 0.097.
	The examples above show that the assumptions mentioned above are reasonable.
\end{Rem}

\subsection{Differential Privacy}
Two databases $D,D' \in \mathcal{D}^n$ differing by one single element are denoted as $D \sim D'$, called \textit{adjacent databases}.

\begin{Def}[Differential Privacy \cite{dwork2014algorithmic}]
	A randomized function $\mathcal{A} : \mathcal{D}^n \rightarrow \mathbb{R}^m$ is ($\epsilon$,$\delta$)-differential privacy (($\epsilon$,$\delta$)-DP) if:
	\begin{equation*}
		\mathbb{P}\left[\mathcal{A}(D) \in S\right] \leq e^\epsilon\mathbb{P}\left[\mathcal{A}(D') \in S\right] + \delta,
	\end{equation*}
	where $S \in$ range($\mathcal{A}$) and $m$ is the dimension of model.
\end{Def}


Differential privacy requires that adjacent datasets $D,D'$ lead to similar distributions on the output of a randomized algorithm $\mathcal{A}$.
This implies that an adversary cannot infer whether an individual participates in the training process because essentially the same conclusions about an individual will be drawn whether or not that individual’s data was used.

In the setting of differentially private machine learning, we consider the following problem model: anyone who can get the model $\theta$ (at each iteration) may be the potential attacker, but they could not get any original data instance in the dataset.
The adversaries attempt to infer the sensitive information of individuals included in the dataset, from the model parameters.
Some kind of attacks, such as membership inference attack, attribute inference attack, memorization attack, can be thwarted by differential privacy \cite{backes2016membership,carlini2019secret,bargav2019evaluating}.

\section{Differentially Private Stochastic Gradient Descent and Utility Bounds}
Considering that SGD naturally fits the condition measuring each data instances differently, before introducing the `Performance Improving' algorithm in detail, we first analyze the excess risk bounds of DP-SGD.

In DP-SGD, the training process at the $t^{th}$ iteration is:
\begin{equation}\label{DPSGD}
	\theta_{t+1}\leftarrow\theta_t-\alpha\left(\nabla_{\theta}\ell(\theta_t,z_t)+b\right),
\end{equation}
where $z_t$ is the chosen data instance at iteration $t$, $\alpha$ is the learning rate, and $b$ is the sampled random noise.

There is a long list of works to analyze the privacy guarantees of DP-SGD.
To the best of our knowledge, the moments accountant method proposed by \cite{abadi2016deep} achieves one of the best results.
It claimed that if the Gaussian random noise $b\sim\mathcal{N}\big(0,\sigma^2I_m\big)$ is injected to (\ref{DPSGD}), the loss function is $G$-Lipschitz, and with
\begin{equation}\label{DPnoise}
	\sigma\geq c\frac{G\sqrt{T\log(1/\delta)}}{n\epsilon}
\end{equation}
for some constant $c$, then the algorithm satisfies $(\epsilon,\delta)$-DP, where $T$ is the total number of training iterations and $n$ is the size of the training dataset.

\subsection{Utility Analysis}
In this section, we analyze the excess emprical risk and the excess population
risk of DP-SGD.
Previous works always discussed the excess empirical risk but seldom considered the excess population risk \cite{song2013stochastic,abadi2016deep,wang2017differentially,zhang2017efficient,wu2017bolt}.
However, the latter is one of the most concerned terms in machine learning because it demonstrates the gap between the private model and the optimal model over the underlying distribution $\mathcal{P}$.

\subsubsection{Excess Empirical Risk}
The excess empirical risk measures the gap between $\theta_{priv}$ and $\theta^*$ over the dataset $D$, where $\theta_{priv}$ denotes the private model.
Before the analysis, we first introduce the Polyak-{\L}ojasiewicz (PL) condition \cite{karimi2016linear}.

\begin{Def}[Polyak-{\L}ojasiewicz condition]
	$L(\theta;D)$ satisfies the Polyak-{\L}ojasiewicz (PL) condition if there exists $\mu>0$ for all $\theta$:
	\begin{equation*}
		\left\Vert\nabla_\theta L(\theta;D)\right\Vert_2^2\geq 2\mu(L(\theta;D)-L^*).
	\end{equation*}
\end{Def}

It is claimed that \cite{karimi2016linear}: Strong Convex $\Rightarrow$ Essential Strong Convexity $\Rightarrow$ Weak Strongly Convexity $\Rightarrow$ Polyak-{\L}ojasiewicz Condition.
PL condition is one of the weakest curvature conditions \cite{li2021improved}, it does not assume the loss function to be convex and it is commonly used in non-convex optimiztion.
Many non-convex models satisfy the condition, including deep (linear) \cite{charles2018stability} and shallow neural networks \cite{lei2021sharper}.

\begin{Rem}
	If $L(\theta;D)$ satisfies the PL condition, then it satisfies the Quadratic Growth (QG) condition \cite{karimi2016linear}, i.e.,
	\begin{equation}\label{QG}
		L(\theta;D)-L(\theta^*;D)\geq\frac{\mu}{2}\|\theta-\theta^*\|_2^2,
	\end{equation}
	where $\theta^*$ denotes the optimal model over dataset $D$.
\end{Rem}

\begin{The}
	Suppose that $\ell(\theta,z)$ is $G$-Lipschitz, $L$-smooth, and satisfies PL condition over $\theta$.
	With learning rate $\alpha=\frac{1}{L}$, $\sigma$ given in (\ref{DPnoise}) to guarantee ($\epsilon,\delta$)-DP, and $T=\mathcal{O}(\log(n))$, then:
	\begin{equation*}
		\mathbb{E}\left[L(\theta_{priv};D)-L^*\right]\leq\mathcal{O}\left(\frac{mG^2\log(1/\delta)\log(n)}{n^2\epsilon^2}\right),
	\end{equation*}
	the expectation is taken over the algorithm and dataset $D$, $m$ denotes the dimensions of the model.
\end{The}

\begin{proof}
	For DP-SGD, we first analyze the bound of $L(\theta_{t+1})-L(\theta_t)$ at iteration $t$.
	Note that the loss function is $L$-smooth (denoted by $L$), with
	$\alpha=\frac{1}{L}$, we have:
	\begin{equation}\label{ITERt}
		\begin{aligned}
			&\mathbb{E}_{b,D}\left[L(\theta_{t+1};D)-L(\theta_t;D)\right] \\
			&\overset{(L)}{\leq}\mathbb{E}_{b,D}\left[\left<\nabla_\theta L(\theta_t;D),
			\theta_{t+1}-\theta_t\right>+\frac{L}{2}\|\theta_{t+1}-\theta_t\|_2^2\right] \\
			&=\mathbb{E}_{b,D}\left[-\frac{1}{L}\left<\nabla_\theta L(\theta_t;D),
			\nabla_\theta\ell(\theta_t,z_t)+b\right>\right]+\mathbb{E}_{b,D}\left[\frac{1}{2L}\|\nabla_\theta\ell(\theta_t,z_t)+b\|_2^2\right] \\
			&\leq-\frac{1}{L}\mathbb{E}_{D}\left[\|\nabla_\theta L(\theta_t;D)\|_2^2\right]+\frac{1}{2L}\left(\mathbb{E}_{D}\left[\|\nabla_\theta
			L(\theta_t;D)\|_2^2\right]+\mathbb{E}\|b\|_2^2\right) \\
			&=-\frac{1}{2L}\mathbb{E}_{D}\left[\|\nabla_\theta
			L(\theta_t;D)\|_2^2\right]+\frac{1}{2L}\mathbb{E}\|b\|_2^2,
		\end{aligned}
	\end{equation}
	the first inequality holds because of $L$-smooth \cite{wang2017differentially} and the second inequality holds because $b$ is zero mean.
	
	Note that fucntion $L(\theta;D)$ satisfies PL condition \cite{csiba2017global}, then:
	\begin{equation*}
		\|\nabla_\theta L(\theta;D)\|_2^2\geq2\mu(L(\theta;D)-L^*).
	\end{equation*}
	
	For random variable $X$ with variance $v(x)$, we have $\mathbb{E}(X^2)=\mathbb{E}^2(X)+v(X)$.
	
	So, with $b\sim\mathcal{N}\left(0,\sigma^2I_m\right)$, inequality (\ref{ITERt}) can be written as:
	\begin{equation*}
		\begin{aligned}
			\mathbb{E}\left[L(\theta_{t+1};D)-L(\theta_t;D)\right]&\leq-\frac{\mu}{L}(\mathbb{E}_{D}\left[L(\theta_t;D)-L^*\right])+\frac{m\sigma^2}{2L}.
		\end{aligned}
	\end{equation*}
	
	As a result,
	\begin{equation*}
		\begin{aligned}
			\mathbb{E}\left[L(\theta_{t+1};D)-L^*\right]&\leq(1-\frac{\mu}{L})(\mathbb{E}_{D}\left[L(\theta_t;D)-L^*\right])+\frac{m\sigma^2}{2L}.
		\end{aligned}
	\end{equation*}
	
	Then, summing over $T$ iterations, we have:
	\begin{equation*}
		\begin{aligned}
			\mathbb{E}_{b,D}[L(\theta_T;D)-L^*]&\leq(1-\frac{\mu}{L})^T(\mathbb{E}_{D}\left[L(\theta_t;D)-L^*\right])+\frac{m\sigma^2}{2L}\left(\sum_{i=0}^{T-1}(1-\frac{\mu}{L})^i\right) \\	
			&\leq(1-\frac{\mu}{L})^T(\mathbb{E}_{D}\left[L(\theta_t;D)-L^*\right])+\frac{m\sigma^2}{2\mu}.
		\end{aligned}
	\end{equation*}
	
	Taking
	$T=\mathcal{O}(\log(n))$ and $\sigma=\mathcal{O}\big(\frac{G\sqrt{T\log(1/\delta)}}{n\epsilon}\big)$ discussed in (\ref{DPnoise}), then with bounded $L\left(\cdot\right)$, we have:
	\begin{equation*}
		\begin{aligned}
			\mathbb{E}[L(\theta_{priv};D)-L^*]&=\mathbb{E}[L(\theta_T;D)-L^*]\leq \mathcal{O}\left(\frac{mG^2\log(1/\delta)\log(n)}{n^2\epsilon^2}\right),
		\end{aligned}
	\end{equation*}
	the expectation is taken over the algorithm and dataset $D$.
	
	The proof is complete.
	
\end{proof}

\begin{Rem}
	Many researchers discussed the excess empirical risk in previous works.
	To the best of our knowledge, one of the best results is given by \cite{wang2017differentially}, in which $T$ is multiplied `rudely' to the noise term when summing the loss over $T$ iterations.
	As a result, the excess empirical risk bound is $\mathcal{O}\big(\frac{mG^2\log(1/\delta)\log^2(n)}{n^2\epsilon^2}\big)$ in \cite{wang2017differentially}.
	However, we solve a geometric sequence when summing the loss, and get a tighter bound in this paper.
	As a result, the excess empirical risk bound is improved by a factor of
	$\log(n)$ overall.
\end{Rem}

\subsubsection{Excess Population Risk}
To get the excess population risk bound, we first analyze the generalization error, which measures the gap between the performance over the underlying distribution and the dataset $D$ of the private model, connecting the population risk with the empirical risk.

\begin{The}
	If the loss function is $G$-Lipschitz, $L$-smooth, and satisfies the PL condition over $\theta$, the generalization error bound of $\theta_{priv}$ satisfies:
	\begin{equation}\label{GE}
		\begin{aligned}
			&\mathbb{E}\left[L_\mathcal{P}(\theta_{priv})-L(\theta_{priv};D)\right] \\
			&\leq\inf_{\tau>0}\left\{\frac{8(\tau+L)}{\mu}\mathbb{E}[L(\theta_{priv};D)-L^*]+\frac{16G^2(\tau+L)}{n^2\mu^2}+\frac{L\mathbb{E}[L(\theta_{priv};D)]}{\tau}\right\},
		\end{aligned}
	\end{equation}
	where the expectation is taken over the algorithm.
\end{The}

Before detailed proof, we introduce the stability theory.

\begin{Def}[Uniform Stability]
	A randomized algorithm $\mathcal{A}$ is $\gamma$-uniform stability if for any neighboring datasets $D\sim D'$,
	\begin{equation*}
		\sup_z\mathbb{E}_\mathcal{A}\left[\ell\left(\theta,z\right)-\ell\left(\theta',z\right)\right]\leq\gamma,
	\end{equation*}
	where $\theta,\theta'$ are derived by datasets $D,D'$, respectively.
\end{Def}

\begin{Lem}[\cite{hardt2016train}]
	If algorithm $\mathcal{A}$ is $\gamma$-uniform stability, then its generalitzation error satisfies:
	\begin{equation*}
		\mathbb{E}_{\mathcal{A}}\left[L_\mathcal{P}(\theta)-L(\theta;D)\right]\leq\gamma.
	\end{equation*}
\end{Lem}

\begin{Lem}[\cite{srebro2010smoothness}]
	If $\ell(\cdot,\cdot)$ is $L$-smooth, then the following self-bounded property holds:
	\begin{equation*}
		\|\nabla_\theta\ell(\theta,z)\|_2^2\leq2L\ell(\theta,z).
	\end{equation*}
\end{Lem}

In the following proof, $\theta^t$ is the model at iteration $t$ and $\theta^t_i$ is the corresponding model derived by its adjacent dataset, in which the $i^{th}$ data instance is the different one.

\begin{proof}
	First, at iteration $t$, if the loss function is $L$-smooth (denoted by $L$) and $G$-Lipschitz (denoted by $G$), then:
	\begin{equation}\label{neighbourt}
		\begin{aligned}
			\ell(\theta^t_i,z)-\ell(\theta^t,z)&\overset{(L)}{\leq}\langle\nabla_{\theta}\ell(\theta^t,z),\theta_i^t-\theta^t\rangle+\frac{L}{2}\left\Vert\theta_i^t-\theta^t\right\Vert_2^2
			\\
			&\leq\left\Vert\nabla_\theta\ell(\theta^t,z)\right\Vert_2\left\Vert\theta_i^t-\theta^t\right\Vert_2+\frac{L}{2}\left\Vert\theta_i^t-\theta^t\right\Vert_2^2
			\\
			&\leq\frac{\left\Vert\nabla_\theta\ell(\theta^t,z)\right\Vert_2^2}{2\tau}+\frac{\tau+L}{2}\left\Vert\theta_i^t-\theta^t\right\Vert_2^2
			\\
			&\leq\frac{L\ell(\theta^t,z)}{\tau}+\frac{\tau+L}{2}\left\Vert\theta_i^t-\theta^t\right\Vert_2^2,
		\end{aligned}
	\end{equation}
	for all $\tau>0$.
	In (\ref{neighbourt}), the second inequality holds because of Cauchy-Schwartz inequality, the third inequality holds because $ab\leq\frac{a^2}{2\tau}+\frac{\tau b^2}{2}$ for $a,b,\tau>0$ and the last inequality holds because of Lemma 2.
	
	Thus, by Lemma 1, we have:
	\begin{equation}\label{generr}
		\begin{aligned}
			\mathbb{E}\left[L_\mathcal{P}(\theta_T)-L(\theta_T;D)\right]&\leq\frac{L\mathbb{E}\left[L(\theta_T;D)\right]}{\tau}+\frac{\tau+L}{2}\mathbb{E}\left[\left\Vert\theta_i^T-\theta^T\right\Vert_2^2\right].
		\end{aligned}
	\end{equation}
	
	By triangle inequality, at iteration $t$,
	\begin{equation*}
		\|\theta_i^t-\theta^t\|_2\leq\|\theta_i^t-\theta_i^*\|_2+\|\theta_i^*-\theta^*\|_2+\|\theta^*-\theta^t\|_2,
	\end{equation*}
	where $\theta_i^*$ denotes the optimal model over the adjacent dataset.
	
	By QG condition discussed in Remark 2 and the symmetry,
	\begin{equation}\label{eq1}
		\|\theta_i^t-\theta_i^*\|_2+\|\theta^*-\theta^t\|_2\leq2\sqrt{2}\sqrt{\frac{L(\theta_t;D)-L(\theta^*;D)}{\mu}}.
	\end{equation}
	
	Then we come to $\|\theta_i^*-\theta^*\|_2$: Due to the Lipschitz property,
	\begin{equation}\label{eq2}
		\begin{aligned}
			L(\theta_i^*;D)-L(\theta^*;D)
			&=\frac{1}{n}\sum_{i=1}^n\ell(\theta_i^*,z_i)-\ell(\theta^*,z_i) \\
			&=\frac{1}{n}\sum_{j\neq i}\ell(\theta_i^*,z_j)-\ell(\theta^*,z_j)+\frac{1}{n}\left(\ell(\theta_i^*,z_i)-\ell(\theta^*,z_i)\right) \\
			&=\frac{1}{n}\sum_{j\neq i}\ell(\theta_i^*,z_j)-\ell(\theta^*,z_j)+\frac{1}{n}\left(\ell(\theta_i^*,z_i)-\ell(\theta^*,z_i)\right) \\
			&\quad+\frac{1}{n}\left(\ell(\theta_i^*,z_i')-\ell(\theta^*,z_i')\right)-\frac{1}{n}\left(\ell(\theta_i^*,z_i')-\ell(\theta^*,z_i')\right) \\
			&=\frac{1}{n}\left(\ell(\theta_i^*,z_i)-\ell(\theta^*,z_i)\right)+\frac{1}{n}\left(\ell(\theta^*,z_i')-\ell(\theta_i^*,z_i')\right) \\
			&\quad+\underbrace{L(\theta_i^*;D')-L(\theta^*;D')}_{\leq0} \\
			&\leq\frac{2G}{n}\|\theta_i^*-\theta^*\|_2,
		\end{aligned}
	\end{equation}
	where the last inequality holds because $\theta_i^*$ is the optimal model over $D'$.
	
	Combining QG condition and (\ref{eq2}), we have:
	\begin{equation*}
		\frac{\mu}{2}\|\theta_i^*-\theta^*\|_2^2\leq L(\theta_i^*;D)-L(\theta^*;D)\leq\frac{2G}{n}\|\theta_i^*-\theta^*\|_2,
	\end{equation*}
	which implies $\|\theta_i^*-\theta^*\|_2\leq\frac{4G}{n\mu}$.
	
	Combining this result with (\ref{eq1}),
	\begin{equation*}
		\|\theta_i^t-\theta^t\|_2\leq2\sqrt{2}\sqrt{\frac{L(\theta_t;D)-L(\theta^*;D)}{\mu}}+\frac{4G}{n\mu}.
	\end{equation*}
	
	Taking this result back to (\ref{generr}), for all $\tau>0$, we have:
	\begin{equation*}
		\begin{aligned}
			\mathbb{E}\left[L_\mathcal{P}(\theta_T)-L(\theta_T;D)\right]&\leq\frac{L\mathbb{E}[\ell(\theta_T,z)]}{\tau}+\frac{\tau+L}{2}\mathbb{E}\left[\left\Vert\theta_i^T-\theta^T\right\Vert_2^2\right] \\
			&\leq(\tau+L)\left(\frac{8}{\mu}\mathbb{E}[L(\theta_T;D)-L^*]+\frac{16G^2}{n^2\mu^2}\right)+\frac{L\mathbb{E}[L(\theta_T;D)]}{\tau},
		\end{aligned}
	\end{equation*}
	which completes the proof.
	
\end{proof}

By Theorem 2, one may observe that the generalization error decreases if the optimization error (the excess empirical risk) is smaller, which is in line with the observation in \cite{hardt2016train,charles2018stability,lei2021sharper}: `optimization helps generalization'.

Now, we give the excess population risk bound.

\begin{The}
	If the loss function is $G$-Lipschitz, $L$-smooth, and satisfies the PL condition over $\theta$, with learning rate $\alpha=\frac{1}{L}$, the excess population risk of $\theta_{priv}$ satisfies:
	\begin{equation*}
		\begin{aligned}
			\mathbb{E}\left[L_\mathcal{P}\left(\theta_{priv}\right)-\min_{\theta}L_\mathcal{P}\left(\theta\right)\right]&\leq(\tau+L)\left(\frac{(8\tau+\mu)}{\mu\tau}\mathbb{E}[L(\theta_{priv};D)-L^*]+\frac{16G^2}{n^2\mu^2}\right) \\
			&\quad+\frac{L}{\tau}\mathbb{E}[L^*].
		\end{aligned}
	\end{equation*}
\end{The}

\begin{proof}
	First, we give some basic notations.
	In the following proof, $\theta^*=\arg\min_\theta L(\theta;D)$ and $\theta^*_\mathcal{P}=\arg\min_\theta L_\mathcal{P}\left(\theta\right)$.
	
	The excess population risk can be divided into:
	\begin{equation}\label{partation}
		\begin{aligned}
			&\mathbb{E}\left[L_\mathcal{P}(\theta_{priv})-\min_{\theta\in\mathcal{C}}L_\mathcal{P}(\theta)\right] \\	
			&=\underbrace{\mathbb{E}\left[L_\mathcal{P}(\theta_{priv})-L(\theta_{priv};D)\right]}_{\rm generalization\quad error}+\underbrace{\mathbb{E}\left[L(\theta_{priv};D)-L(\theta^*;D)\right]}_{\rm excess\quad empirical\quad risk}+\underbrace{\mathbb{E}\left[L(\theta^*;D)-\min_{\theta\in\mathcal{C}}L_\mathcal{P}(\theta)\right]}_{C}.
		\end{aligned}
	\end{equation}
	
	In (\ref{partation}), the first and second parts on the right side of the equation are analyzed before.
	And the only thing left is part $C$, which can be written as:
	\begin{equation*}
		L(\theta^*;D)-\min_{\theta\in\mathcal{C}}L_\mathcal{P}(\theta)=L(\theta^*;D)-\mathbb{E}_D\left[L(\theta_\mathcal{P}^*;D)\right]\leq0.
	\end{equation*}
	The last inequality holds because $\theta_\mathcal{P}^*$ is independent of dataset $D$, and $\theta^*=\arg\min_{\theta\in\mathcal{C}}L(\theta;D)$.
	
	As a result, the excess population risk satisfies:
	\begin{equation*}
		\begin{aligned}
			&\mathbb{E}\big[L_\mathcal{P}(\theta_{priv})-\min_{\theta\in\mathcal{C}}L_\mathcal{P}(\theta)\big]\leq\mathbb{E}\left[L_\mathcal{P}(\theta_{priv})-L(\theta_{priv};D)\right]+\mathbb{E}\left[L(\theta_{priv};D)-L(\theta^*;D)\right].
		\end{aligned}
	\end{equation*}
	
	Taking the results given in Theorems 1 and 2, we have:
	\begin{equation*}
		\begin{aligned}
			&\mathbb{E}\left[L_\mathcal{P}(\theta_{priv})-\min_{\theta\in\mathcal{C}}L_\mathcal{P}(\theta)\right]\leq(\tau+L)\left(\frac{(8\tau+\mu)}{\mu\tau}\mathbb{E}[L(\theta_T;D)-L^*]+\frac{16G^2}{n^2\mu^2}\right)+\frac{L}{\tau}L^*,
		\end{aligned}
	\end{equation*}
	which completes the proof.
\end{proof}

\begin{Rem}
	Combining the result given in Theorem 1, if $\mathbb{E}[L^*]=\mathcal{O}(1/n)$, taking $T=\mathcal{O}(\log(n))$, if we ignore constants and $\log(\cdot)$ terms, then for all $\tau>0$, the excess population risk bound comes to:
	\begin{equation*}
		\mathcal{O}\left(\frac{m}{\tau n^2\epsilon^2}+\frac{\tau m}{n^2\epsilon^2}+\frac{1}{\tau n}\right).
	\end{equation*}
	If $\tau=\mathcal{O}(1)$,
	\begin{equation*}
		\mathbb{E}\left[L_\mathcal{P}\left(\theta_{priv}\right)-\min_{\theta}L_\mathcal{P}\left(\theta\right)\right]=\mathcal{O}\left(\frac{m}{n^2\epsilon^2}+\frac{1}{n}\right).
	\end{equation*}
	If $\tau=\mathcal{O}(\sqrt{n}\epsilon/\sqrt{m})$,
	\begin{equation*}
		\mathbb{E}\left[L_\mathcal{P}\left(\theta_{priv}\right)-\min_{\theta}L_\mathcal{P}\left(\theta\right)\right]=\mathcal{O}\left(\frac{\sqrt{m}}{n^{1.5}\epsilon}+\frac{m^{1.5}}{n^{2.5}\epsilon^3}\right).
	\end{equation*}
	Thus, we get the upper bound of the excess population risk:
	\begin{equation*}
		\mathcal{O}\left(\min\left\{\frac{m}{n^2\epsilon^2}+\frac{1}{n},\frac{\sqrt{m}}{n^{1.5}\epsilon}+\frac{m^{1.5}}{n^{2.5}\epsilon^3}\right\}\right).
	\end{equation*}
\end{Rem}

In Theorem 3, to get the better result, we assume $\mathbb{E}[L^*]=\mathcal{O}(1/n)$.
It is a small value because $L^*$ is the optimal value over the whole dataset.
Besides, it is common to assume the minimal population risk $\mathbb{E}[\min L_\mathcal{P}(\theta)]\leq\mathcal{O}(1/n)$ \cite{lei2020fine,zhang2019stochastic,liu2018fast,zhang2017empirical,srebro2010optimistic}.
Moreover, under expectation, considering $\theta_\mathcal{P}^*=\arg\min L_\mathcal{P}(\theta)$ is independent of dataset, so $\mathbb{E}[L^*]\leq\mathbb{E}[\min L_\mathcal{P}(\theta)]$ \cite{lei2020sharper}.
Thus, the assumption is reasonable.

All the theorems given above only assume that the loss function $\ell(\cdot)$ is $G$-Lipschitz, $L$-smooth and satisfies PL inequality, without convex assumption.
So the results are general and can be applied to some of the non-convex conditions.

\section{Performance Improving Differentially Private Stochastic Gradient Descent}

Motivated by the definition of DP, we focus on the contributions made by data instances on the final model.
In particular, if the effects caused by a data instance $z$ on the final machine learning model is so little that the attacker cannot realize it (less than $e^{\epsilon}$), there is no need to add noise to $z$.
Now, only one problem is left: How to measure the impact of the data instances on the model?
A classic technique, Influence Function (IF), gives us some inspirations.

\subsection{Influence Function}
The contribution of data instance $z$ is naturally defined as $\theta_{-z}^*-\theta^*$, where $\theta_{-z}^*=\arg\min_{\theta}\sum_{z_i\neq z}\ell(\theta,z_i)$.
To measure the gap between them, a straight method is to train two models: $\theta^{*}$, $\theta^*_{-z}$.
However, retraining a model for each data instance $z$ is prohibitively slow.
To solve the problem, influence function introduced by
\cite{koh2017understanding} measures the contributions on the machine learning
model made by data instances:
\begin{equation}\label{IF}
	c_z\coloneqq-\frac{1}{n}\left(-H_{\theta^*}^{-1}\nabla_\theta\ell\left(\theta^*,z\right)\right)\approx\theta_{-z}^*-\theta^*,
\end{equation}
where $H_{\theta^*}=\frac{1}{n}\sum_{i=1}^{n}\nabla_\theta^2\ell(\theta^*,z_i)$, assumed positive definite.
Via (\ref{IF}), we can measure how the model changes if we `drop' one data instance, naturally in line with the definition of DP.

\subsection{Error Analysis}
The influence function $c_z$ is got by Taylor expansion \cite{linnainmaa1976taylor}, in which Taylor remainders lead an approximation error.
However, \cite{koh2017understanding} only gives an approximation via IF, but not discusses the corresponding error.
To fill the gap, we analyze the approximation error in this section.

\begin{The}
	If the loss function $\ell(\theta,z)$ is $G$-Lipschitz, $L$-smooth, $C$-Hessian Lipschitz over $\theta$, and $\left\Vert H_{\theta^*}\right\Vert_2\geq\zeta$, then the approximation error satisfies:
	\begin{equation*}
		E\coloneqq\|(\theta_{-z}^*-\theta^*)-c_z\|_2\leq\frac{1}{\zeta^2n^2}\left(2LG+\frac{CG^2}{\zeta}\right),
	\end{equation*}
	where $c_z$ is the approximation given in (\ref{IF}).
\end{The}

\begin{proof}
	Before detailed proof, we first define some new notations.
	
	With dataset $D$, and weight vector $\boldsymbol{w}$, we define:
	\begin{equation*}	
		\hat{\theta}(\boldsymbol{w};D)=\arg\min_{\theta\in\mathcal{C}}
		F(\boldsymbol{w},\theta,D)\coloneqq\frac{1}{n}\sum_{i=1}^{n}w_i\ell\left(\theta,z_i\right),
	\end{equation*}
	where $z_i\in D$ and $\boldsymbol{w}\in\{0,1\}^n$ is a vector of sample weights with $w_i=1$ if we use $z_i$ when training, otherwise $0$.
	
	In this way, $\boldsymbol{w}=\boldsymbol{1}$ means all of the data instances are used and $\boldsymbol{w}=\boldsymbol{1}_{\setminus r}$ means expect for $i=r$, $w_i=1$.
	
	Note that $\hat{\theta}(\boldsymbol{w};D)$ is the optimal solution of function $F(\boldsymbol{w},\theta,D)$, so:
	\begin{equation*}
		\nabla_\theta
		F(\boldsymbol{w},\hat{\theta}(\boldsymbol{w}),D)=\frac{1}{n}\sum_{i=1}^{n}w_i\nabla_\theta\ell\left(\hat{\theta}(\boldsymbol{w}),z_i\right)=0.
	\end{equation*}
	
	Taking the first derivative on both sides over $w_i$ yields:
	\begin{equation}\label{firstd}
		\frac{1}{n}\nabla_\theta\ell\left(\hat{\theta}(\boldsymbol{w}),z_i\right)+\frac{1}{n}\sum_{j=1}^{n}w_j\nabla_{\theta}^2\ell\left(\hat{\theta}(\boldsymbol{w}),z_j\right)\frac{\mathrm{d}\hat{\theta}(\boldsymbol{w})}{\mathrm{d}w_i}=0.
	\end{equation}
	
	Taking the first derivative on both sides over $w_j$ yields:
	\begin{equation*}
		\begin{aligned}
			0&=\frac{1}{n}\nabla_\theta^2\ell\left(\hat{\theta}(\boldsymbol{w}),z_i\right)\frac{\mathrm{d}\hat{\theta}(\boldsymbol{w})}{\mathrm{d}w_j}+\frac{1}{n}\nabla_\theta^2\ell\left(\hat{\theta}(\boldsymbol{w}),z_j\right)\frac{\mathrm{d}\hat{\theta}(\boldsymbol{w})}{\mathrm{d}w_i}
			\\
			&\quad+\frac{1}{n}\sum_{k=1}^{n}w_k\nabla_{\theta}^3\ell\left(\hat{\theta}(\boldsymbol{w}),z_k\right)\frac{\mathrm{d}\hat{\theta}(\boldsymbol{w})}{\mathrm{d}w_i}\left[\frac{\mathrm{d}\hat{\theta}(\boldsymbol{w})}{\mathrm{d}w_j}\right]^T
			\\
			&\quad+\frac{1}{n}\sum_{k=1}^{n}w_k\nabla_{\theta}^2\ell\left(\hat{\theta}(\boldsymbol{w}),z_k\right)\frac{\mathrm{d}\hat{\theta}(\boldsymbol{w})}{\mathrm{d}w_j\mathrm{d}w_i}.
		\end{aligned}
	\end{equation*}
	
	Via (\ref{firstd}), we have:
	\begin{equation}\label{firstordresult}
		\frac{\mathrm{d}\hat{\theta}(\boldsymbol{w})}{\mathrm{d}w_i}=-H_{\hat{\theta}(\boldsymbol{w})}^{-1}\left(\frac{1}{n}\nabla_\theta\ell\left(\hat{\theta}(\boldsymbol{w}),z_i\right)\right),
	\end{equation}
	where $H_{\hat{\theta}(\boldsymbol{w})}=\frac{1}{n}\sum_{j=1}^{n}w_j\nabla_{\theta}^2\ell\left(\hat{\theta}(\boldsymbol{w}),z_j\right)$.
	
	Then, with:
	\begin{equation*}
		\begin{aligned}
			H^i=\frac{1}{n}\nabla_\theta^2\ell\left(\hat{\theta}(\boldsymbol{w}),z_i\right),\quad T_{\boldsymbol{w}}=\frac{1}{n}\sum_{k=1}^{n}w_k\nabla_{\theta}^3\ell\left(\hat{\theta}(\boldsymbol{w}),z_k\right),
		\end{aligned}
	\end{equation*}
	we have:
	\begin{equation*}
		\begin{aligned}
			-H_{\hat{\theta}(\boldsymbol{w})}\frac{\mathrm{d}\hat{\theta}(\boldsymbol{w})}{\mathrm{d}w_j\mathrm{d}w_i}&=H^i\frac{\mathrm{d}\hat{\theta}(\boldsymbol{w})}{\mathrm{d}w_j}+H^j\frac{\mathrm{d}\hat{\theta}(\boldsymbol{w})}{\mathrm{d}w_i}+T_{\boldsymbol{w}}\frac{\mathrm{d}\hat{\theta}(\boldsymbol{w})}{\mathrm{d}w_i}\left[\frac{\mathrm{d}\hat{\theta}(\boldsymbol{w})}{\mathrm{d}w_j}\right]^T.
		\end{aligned}
	\end{equation*}
	
	Taking the result given in (\ref{firstordresult}) to the equation above, and noting that the loss function is $G$-Lipschitz, $L$-smooth, and $C$-Hessian Lipschitz, if $\left\Vert H_{\hat{\theta}(\boldsymbol{w})}\right\Vert_2\geq\zeta$, we have:
	\begin{equation*}
		\left\Vert\frac{\mathrm{d}\hat{\theta}(\boldsymbol{w})}{\mathrm{d}w_j\mathrm{d}w_i}\right\Vert_2\leq\frac{1}{\zeta}\left(\frac{2LG}{n^2\zeta}+\frac{CG^2}{n^2\zeta^2}\right).
	\end{equation*}
	
	Now, we consider the following condition:
	\begin{equation*}
		\begin{aligned}
			\hat{\theta}(\boldsymbol{1};D)&=\arg\min_{\theta\in\mathcal{C}}F(\boldsymbol{1},\theta,D)\coloneqq\frac{1}{n}\sum_{i=1}^{n}\ell\left(\theta,z_i\right), \\
			\hat{\theta}(\boldsymbol{1}_{\setminus r};D)&=\arg\min_{\theta\in\mathcal{C}}F(\boldsymbol{1}_{\setminus r},\theta,D)\coloneqq\frac{1}{n-1}\sum_{i=1}^{n}w_i\ell\left(\theta,z_i\right),
		\end{aligned}
	\end{equation*}
	which yields $\theta_{-z_r}^*-\theta^*=\hat{\theta}(\boldsymbol{1}_{\setminus r})-\hat{\theta}(\boldsymbol{1})$.
	
	By Taylor expansion, we have:
	\begin{equation}\label{taylor}
		\begin{aligned}
			\hat{\theta}(\boldsymbol{1}_{\setminus r})&=\hat{\theta}(\boldsymbol{1})+\frac{\mathrm{d}\hat{\theta}(\boldsymbol{1})}{\mathrm{d}\boldsymbol{w}^T}\left(\boldsymbol{1}_{\setminus r}-\boldsymbol{1}\right)+\underbrace{\sum_{k,h=1}^{n}\left[\boldsymbol{1}_{\setminus r}-\boldsymbol{1}\right]_k\left[\boldsymbol{1}_{\setminus r}-\boldsymbol{1}\right]_h\frac{\mathrm{d}\hat{\theta}(\boldsymbol{w'})}{\mathrm{d}w_k\mathrm{d}w_h}}_{R},
		\end{aligned}
	\end{equation}
	$\boldsymbol{w'}\in\left[\boldsymbol{1}_{\setminus r},\boldsymbol{1}\right]$, $\left[\boldsymbol{1}_{\setminus r}-\boldsymbol{1}\right]_j=0$ if $w_j=0$, otherwise $-1$.
	
	Note that $\theta^*=\hat{\theta}(\boldsymbol{1})$, $-\frac{\mathrm{d}\hat{\theta}(\boldsymbol{1})}{\mathrm{d}w_i}$ is the same as
	the IF of $z_i$.
	
	In (\ref{taylor}),
	\begin{equation*}
		\frac{\mathrm{d}\hat{\theta}(\boldsymbol{1})}{\mathrm{d}\boldsymbol{w}^T}\left(\boldsymbol{1}_{\setminus r}-\boldsymbol{1}\right)=-\frac{\mathrm{d}\hat{\theta}(\boldsymbol{1})}{\mathrm{d}w_r}.
	\end{equation*}
	
	So the approximation error between the IF and $\theta_{-z_r}^*-\theta^*$ is exactly term $R$ in (\ref{taylor}), i.e. $E=\left\Vert R\right\Vert_2$.
	
	In Algorithm 1, $\left\Vert H_{\theta^*}\right\Vert_2\geq\zeta$, and under the condition that $\boldsymbol{w}=\boldsymbol{1}$, $\left[\boldsymbol{1}_{\setminus r}-\boldsymbol{1}\right]_j=-1$ only when $j=r$, so we have:
	\begin{equation*}
		E=\left\Vert
		R\right\Vert_2\leq\left\Vert\frac{\mathrm{d}\hat{\theta}(\boldsymbol{w'})}{\mathrm{d}w_k\mathrm{d}w_h}\right\Vert_2\leq\frac{1}{\zeta}\left(\frac{2LG}{n^2\zeta}+\frac{CG^2}{n^2\zeta^2}\right).
	\end{equation*}
	
	The proof is complete.
	
\end{proof}

Theorem 4 gives a $\mathcal{O}\big(1/n^2\big)$ approximation error when applying $c_z$, which means that $c_z$ is precise.

\begin{Rem}
	In Theorem 4, we assume that $\left\Vert H_{\theta^*}\right\Vert_2\geq\zeta$.
	Because most of the algorithms are regularized, so the assumption is easy to hold. And if $H_{\theta^*}$ does not satisfy the assumption, we can add a small $\zeta>0$ to the diagonal elements of $H_{\theta^*}$.
\end{Rem}

\begin{algorithm}[tb]
	\caption{Performance Improving DP-SGD}
	\label{alg2}
	\begin{algorithmic}[1]
		\Require dataset $D$, learning rate $\alpha$, local iteration rounds
		$T_{local}$, global update rounds $R$
		\Function {PIDP-SGD}{$D,\alpha,T_{local},R$}
		\State Initialize $\theta^{(g)}_0, \theta^{(l)}_0 \leftarrow
		\widetilde{\theta}$.
		\State \textbf{for} $r=0$ to $R-1$ \textbf{do}
		\State \quad Get $H_{\widetilde{\theta}^{(g)}_r}$ and compute its inverse.
		\State \quad\textbf{for} $t=0$ to $T_{local}-1$ \textbf{do}
		\State \qquad Choose data instance $z_t$ randomly.
		\State \qquad Get contribution of $z_t$: $c_t^{(o)}=c_{z_t}+E$, via
		$H_{\widetilde{\theta}^{(g)}_r}$.
		\State \qquad Sample $b^{(c)}\sim \mathcal{N}(0,\sigma_{(c)}^2I_m)$.
		\State \qquad $c_t=2c_t^{(o)}+b^{(c)}$, if there exists any $i$ that $|[c_t]_i|\geq\frac{2re^{\epsilon_1}\delta_1}{e^{\epsilon_1}-1}$, jump to line 13; otherwise, jump to line 10.
		\State \qquad\textbf{if}
		$\ln\big(\frac{2re^{\epsilon_1}\delta_1}{2re^{\epsilon_1}\delta_1-\left(e^{\epsilon_1}-1\right)sign(\left[c_t\right]_i)\left[c_t\right]_i}\big)\leq2\epsilon_1$ for all $i\in\left[1,m\right]$,
		\textbf{then}
		\State \qquad\quad
		$\theta^{(l)}_{t+1} \leftarrow
		\theta^{(l)}_t-\alpha\nabla_{\theta}\ell(\theta^{(l)}_t,z_t).$
		\State \qquad\textbf{else}
		\State \qquad Sample $b\sim\mathcal{N}(0,\sigma^2I_m)$,
		\State \qquad
		$\theta^{(l)}_{t+1} \leftarrow
		\theta^{(l)}_t-\alpha\big(\nabla_{\theta}\ell(\theta^{(l)}_t,z_t)+b\big)$.
		\State \qquad\textbf{endif}
		\State \quad \textbf{endfor}
		\State
		$\theta^{(g)}_{r+1}=\theta^{(l)}_{T_{local}}.$
		\State \textbf{endfor}
		\State return $\theta_{priv}=\theta^{(g)}_R$.
		\EndFunction
	\end{algorithmic}
\end{algorithm}

\subsection{Performance Improving DP-SGD}

To measure contributions made by training data instances, we need a (near) optimal model to calculate $c_z$.
Then, we set a threshold: $e^\epsilon$ for adding noise, based on the following observation: the appearance (or absence) of some data affects the model so little that attackers cannot infer anything from them.
Changing those data instances cannot threaten ($\epsilon,\delta$)-DP of the model.
So, we calculate the `contribution' of $z$ by IF, only add noise to the data who contributes more than $e^\epsilon$.

Details of the Performance Improving algorithm are given in Algorithm 1\footnote{In Algorithm \ref{alg2}, $sign(\cdot)$ is the signum function, and $r$ is the radius of the bounded parameter space.}.
Different from traditional DP-SGD algorithm, Algorithm \ref{alg2} applies a decision process before gradient descent (lines 9 and 10), to decide whether to add random noise or not.
If the effect made by the chosen $z_t$ is no more than $e^\epsilon$, SGD runs; otherwise, we sample Gaussian noise $b$ and run DP-SGD.
In other words, lines 9 and 10 connects the value of IF with the privacy loss of DP, details can be found in the proof of Theorem 5.
Meanwhile, we notice that except for the training process, the contribution calculating process may also disclose the sensitive information.
So we add noise to the contirbution $c_t^{(o)}$ to guarantee the claimed differential privacy (line 9).
Noting that the contribution given by (\ref{IF}) is based on Taylor expansion, causing an approximation error due to Taylor remaineders (discussed in Section
\uppercase\expandafter{\romannumeral5}.$B$), we fix it by adding $E$ to the contribution $c_t$ in line 7.


It is easy to follow that if the privacy budget $\epsilon$ is higher, the constraint of adding noise is looser, which means that fewer data instances meet the noise.
As a result, the performance of PIDP-SGD will be better if $\epsilon$ is higher.
If a new data instance $z_{new}$ comes\footnote{The case that old data instance leaves is similar.}, we can directly compute the contribution of $z_{new}$ (denoted by $c_{new}$).
By combining it with current model $\theta_{priv}$, we can get whether $z_{new}$ would change the model significantly.
In this way, if we want to train a new model with $c_{new}$, we can choose whether to add noise or not.
Besides, influence function corresponding to Mini-Batch gradient descent method can be easily extended via the analysis given in \cite{zhu2021general}.
In particular, it is shown that the variance of the noise injected into the gradient of the `batch' is related to the batch size \cite{abadi2016deep} ($B^2/n^2$ w.r.t the sampling probability), so we can modify the sampling to $B_{IF}^2/n^2$, where $B_{IF}$ means the number of `important' data instances in the batch.
We only use SGD as an example to prove that IF can be applied to gradient descent methods to improve the model performance and do not focus on the variants (such as Mini-Batch method) here.
And the method given in this paper will inspire other researchers to apply it to corresponding fields.

\begin{Rem}
	In Algorithm \ref{alg2}, $c_t=2c_t^{(o)}+b^{(c)}$.
	The coefficient $2$ is because in DP, besides inserting (or deleting) a data instance to (or from) the dataset, exchanging is also allowed.
\end{Rem}


\begin{Rem}
	In Algorithm \ref{alg2}, we consider the differential privacy of the whole algorithm.
	Because the Hessian may change during the training process, so the sequence of the data samples may affect the noise adding process.
	For example, when we first choose $z$, the contribution may be significant and noise is injected.
	However, with the changes of the Hessian, when we choose $z$ later, the contribution may decrease we do not need to add noise when training with $z$.
	Under these circumstances, the sequence of the data flow indeed affects the noise adding process, but the DP property still holds, because for the new model, data $z$ is not so significant as before.
\end{Rem}

\begin{Rem}
	The time complexity of Algorithm \ref{alg2} is $\mathcal{O}\big(Rnm^2+RT_{local}m\big)$.
	Under the case $T_{local}=1$, it becomes $\mathcal{O}\big(Rnm^2\big)$, where $R$ is the total number of iterations.
	Fortunately, an efficient approach to calculate the Influence Function was given in \cite{koh2017understanding}, and the time complexity can be reduced to $\mathcal{O}\big(Rnm\big)$.
	For some other previous performance improving method, the time complexity also increases, we show DP-LSSGD as an example here, whose time complexity is $\mathcal{O}\big(Rm^2\big)$.
	Under the cases $n>m$, our time complexity is larger than DP-LSSGD, and under high dimension cases, when $n\leq m$, our time complexity is better.
	However, both the theoretical and the experimental results of our method is much better than DP-LSSGD (shown in TABLE \uppercase\expandafter{\romannumeral1} and Figures~\ref{Accuracy} and ~\ref{Optimality gap}).
	So under low dimension conditions, the sacrifice on time complexity is a trade-off against the model performance; and under high dimension conditions, our method is much better on both the model performance and the time complexity.
\end{Rem}

\subsection{Privacy Guarantees}

\begin{The}
	In Algorithm 1, for $\delta_1,\delta_2>0$ and $\epsilon_1,\epsilon_2>0$, if $\ell(\theta,z)$ is $G$-Lipschitz over $\theta$, with
	\begin{equation*}
		\sigma\geq c\frac{G\sqrt{T\log(1/\delta_1)}}{n\epsilon_1},\quad\sigma_{(c)}\geq c' \frac{GR\sqrt{\log(1.25R/\delta_2)}}{n\zeta\epsilon_2},
	\end{equation*}
	where $T=T_{local}*R$.
	It is ($\epsilon_1+\epsilon_2,\delta_1+\delta_2$)-DP for some constants $c,c'$.
\end{The}

Before the detailed proof, we first revisit the Gaussian mechanism.

\begin{Def}[$\ell_2$-sensitivity \cite{dwork2014algorithmic}]
	For $D\sim D'$, the $\ell_2$-sensitivity of a function $f$ is defined as:
	\begin{equation*}
		S(f)=\max\|f(D)-f(D')\|_2.
	\end{equation*}
\end{Def}

The $\ell_2$-sensitivity captures the magnitude by which a single individual’s data can change $f$ in the worst case.

\begin{Lem}
	By Gaussian mechanism proposed in \cite{dwork2014algorithmic}, ($\epsilon$,$\delta$)-DP can be guaranteed if random noise $b\sim \mathcal{N}(0,\sigma^2)$ is added to a query, where $\sigma\geq c\frac{S(f)}{\epsilon}$ and $c>\sqrt{2\log(1.25/\delta)}$.
\end{Lem}

\begin{proof}
	We first analyze the privacy when computing the contributions $c_t$ made by data instances.
	
	Without privacy consideration, we have:
	\begin{equation*}
		c_t=\widetilde{\theta}_{-z_t}-\theta^{(g)}_r=-\frac{1}{n}\left(-H_{\widetilde{\theta}}^{-1}\nabla_{\theta}\ell(\theta^{(g)}_r,z_t)\right),
	\end{equation*}
	where
	$H_{\widetilde{\theta}}=\frac{1}{n}\sum_{i=1}^{n}\nabla_\theta^2\ell\left(\widetilde{\theta},z_i\right)$.
	
	As a result, on adjacent databases, we have:
	\begin{equation*}
		\begin{aligned}
			S(c_t)&=\|c_t(D)-c_t(D')\|_2=\frac{1}{n}\left\|H_{\widetilde{\theta}}^{-1}\left(\nabla\ell(\theta^{(g)}_r,z_t)-\nabla\ell(\theta^{(g)}_r,z_t')\right)\right\|_2.
		\end{aligned}
	\end{equation*}
	
	If $\ell(\cdot)$ is $G$-Lipschitz over $\theta$, and $\left\Vert H_{\widetilde{\theta}}\right\Vert_2\geq\zeta$, we have:
	\begin{equation*}
		S(c_t)\leq\frac{2G}{n\zeta}.
	\end{equation*}
	
	As shown in Algorithm \ref{alg2}, the contribution $c_{z_t}$ for data instance $z_t$ is fixed during $T_{local}$ iterations and the fixed value only derives the privacy loss once.
	So the contribution made by $z_t$ changes at most $R$ times during the whole $T$ iterations, and we only need to sum the privacy loss at most $R$ times.
	
	Therefore, by Lemma 3, when computing $c_t$, if the added Gaussian random noise $b^{(c)}\sim\mathcal{N}\left(0,\sigma^2\right)$ satisfies:
	\begin{equation*}
		\sigma>\frac{2\sqrt{2}\sqrt{\log(1.25R/\delta_2)}GR}{n\zeta\epsilon_2},
	\end{equation*}
	($\frac{\epsilon_2}{R}$, $\frac{\delta_2}{R}$)-DP at iteration $t$ is
	guaranteed by Gaussian mechanism.
	So, with $c'=2.83$, $c_t$ calculation is ($\epsilon_2$, $\delta_2$)-DP overall.

	Then, we consider the privacy which is naturally guaranteed.
	With the definition of DP (Definition 4), we have:
	\begin{equation*}
		\mathbb{P}\left[\mathcal{A}(D)\in S\right]\leq e^{\epsilon}\mathbb{P}\left[\mathcal{A}(D')\in S\right]+\delta.
	\end{equation*}
	
	Via the discussion in \cite{dwork2006differential}, if all $m$ dimensions of model $\mathcal{A}(D)$ are ($\epsilon,\delta$)-DP, then ($\epsilon,\delta$)-DP of $\mathcal{A}(D)$ is guaranteed.
	So, in the following, we consider each dimension of $\mathcal{A}(D)$.
	
	($\epsilon,\delta$)-DP naturally holds if the following inequality holds:
	\begin{equation*}
		1-\frac{\delta}{\mathbb{P}\left[\left[\mathcal{A}(D)\right]_i\in S\right]}\leq e^{\epsilon}\frac{\mathbb{P}\left[\left[\mathcal{A}(D')\right]_i\in S\right]}{\mathbb{P}\left[\left[\mathcal{A}(D)\right]_i\in S\right]}.
	\end{equation*}
	
	For simpicity, we set $\mathbb{P}\left[\left[\mathcal{A}(D)\right]_i\in S\right]=\frac{\kappa\delta}{\kappa-1}$ ($\kappa>1$), then:
	\begin{equation*}
		\frac{1}{\kappa}\leq e^{\epsilon}\frac{\mathbb{P}\left[\left[\mathcal{A}(D')\right]_i\in S\right]}{\mathbb{P}\left[\left[\mathcal{A}(D)\right]_i\in S\right]},
	\end{equation*}
	which can be written as:
	\begin{equation}\label{DPi}
		\ln\left(\frac{\mathbb{P}\left[\left[\mathcal{A}(D)\right]_i\in S\right]}{\mathbb{P}\left[\left[\mathcal{A}(D')\right]_i\in S\right]}\right)\leq\epsilon+\ln(\kappa).
	\end{equation}
	
	Considering that the model paramaters are uniformly distributed in a bounded parameter space, whose radius is $r$, i.e., $\left[\mathcal{A}(D')\right]_i\in[-r,r]$, and noting that $\mathbb{E}\left[\mathcal{A}(D')\right]=\mathcal{A}(D)$, where the expectation is taken over the single different $z$ between $D$ and $D'$, then $\mathbb{P}\left[\left[\mathcal{A}(D)\right]_i\in S\right]$ is located in $\big[\frac{-r\kappa\delta}{(\kappa-1)},\frac{r\kappa\delta}{(\kappa-1)}\big]$.
	
	With the largest fluctuation $c_t$, then for the $i^{th}$ element of $\mathcal{A}(D)$ and $\mathcal{A}(D')$,
	\begin{equation*}
		\left[\mathcal{A}(D')-\mathcal{A}(D)\right]_i=\left[\mathcal{A}(D')\right]_i-\left[\mathcal{A}(D)\right]_i\leq\left[c_t\right]_i,
	\end{equation*}
	where $[\cdot]_i$ represents the $i^{th}$ element of a vector.
	
	As a result, if $\left[c_t\right]_i\geq0$, then $\big[[c_t]_i-\frac{r\kappa\delta}{(\kappa-1)},\frac{r\kappa\delta}{(\kappa-1)}\big]$ is the area that $\mathbb{P}\left[\left[\mathcal{A}(D')\right]_i\in S\right]$; otherwise, $\big[\frac{-r\kappa\delta}{(\kappa-1)},[c_t]_i+\frac{r\kappa\delta}{(\kappa-1)}\big]$ is the area that $\mathbb{P}\left[\left[\mathcal{A}(D')\right]_i\in S\right]$.
	
	Thus, we have:
	\begin{equation*}
		\mathbb{P}\left[\left[\mathcal{A}(D')\right]_i\in S\right]\leq
		\left\{
		\begin{aligned}
			&\frac{\kappa\delta}{(\kappa-1)}-\frac{\left[c_t\right]_i}{2r}, \quad \left[c_t\right]_i\geq0, \\
			&\frac{\kappa\delta}{(\kappa-1)}+\frac{\left[c_t\right]_i}{2r}, \quad \left[c_t\right]_i<0.
		\end{aligned}
		\right.
	\end{equation*}
	
	If $|[c_t]_i|\geq\frac{2r\kappa\delta}{(\kappa-1)}$, then $\mathbb{P}\left[\left[\mathcal{A}(D)\right]_i\in S\cap\left[\mathcal{A}(D')\right]_i\in S\right]=0$ so $z_t$ has significant contribution, we directly jump to line 13 as shown in Algorithm 1.
	If $|[c_t]_i|<\frac{2r\kappa\delta}{(\kappa-1)}$ for all $i$, then by (\ref{DPi}), ($\epsilon_1,\delta_1$)-DP is guaranteed if for all $m$ dimensions, the following inequality holds:
	\begin{equation*}
		\ln\left(\frac{2r\kappa\delta_1}{2r\kappa\delta_1-(\kappa-1)sign(\left[c_t\right]_i)\left[c_t\right]_i}\right)\leq\epsilon_1+\ln(\kappa),
	\end{equation*}
	where $sign(\cdot)$ is the signum function and $1\leq i\leq m$.
	
	We set $\kappa=e^{\epsilon_1}$, then via Algorithm 1, ($\epsilon_1,\delta_1$)-DP is naturally guaranteed in lines 10 and 11.
	
	Finally, we consider the privacy when training the model.
	
	For the data instances whose contributions are not so much, $(\epsilon_1,\delta_1)$-DP is naturally guaranteed, as analyzed above.
	And if the contribution is large, via moments accountant method, if $b\sim\mathcal{N}\big(0,\sigma^2I_m\big)$ and $\sigma=\mathcal{O}\big(\frac{G\sqrt{T\log(1/\delta_1)}}{n\epsilon_1}\big)$, then ($\epsilon_1,\delta_1$)-DP can be guaranteed when training.
	
	Then, by composition theorems \cite{dwork2014algorithmic}, Algorithm 1 is ($\epsilon_1+\epsilon_2$, $\delta_1+\delta_2$)-DP overall, which completes the proof.
	
\end{proof}

Theorem 5 shows that the privacy of Algorithm \ref{alg2} consists of two parts: (1) computing the contribution $c_t$ and (2) training the model by DP-SGD.
Specifically, $\epsilon_1,\delta_1$ ($\sigma$) are for the privacy when training the model and $\epsilon_2,\delta_2$ ($\sigma_{(c)}$) are for the privacy when computing $c_t$.

\begin{Rem}
	In the proof, we analyze the DP of the SGD process by the moments accountant method and for DP of computing $c_t$, the analysis is based on the $\ell_2$-sensitivity and Gaussian mechanism.
	The reason is that for SGD, tracking the privacy loss during $T$ iterations is an important part and the moments accountant method naturally solves the problem.
	However, computing $c_t$ is only for the decision process and during $T_{local}$ iterations, $c_{z_t}$ for a certain data instance $z_t$ is fixed and only calculated once, so it derives corresponding privacy loss only once.
	Thus, there is no need to track the privacy loss over the whole training iterations.
	As a result, we discuss the privacy leakage brought by $c_t$ from the point of view: $\ell_2$-sensitivity and Gaussian mechanism.
\end{Rem}

In Algorithm 1, noise is only added when training with a partition of data instances, which leads better excess risk bounds.
In the following, we suppose that there are $k$ data instances affect the model significantly and measure the improvement brought by our proposed `Performance Improving' algorithm.

\begin{table*}
	\center
	\caption{Comparisons on excess risk bounds between our method and other methods.}
	\label{table2}
	\begin{threeparttable}
		\begin{tabular}{ccccccc}
			\hline
			& $G$ & $L$ & S.C. & C & PL & EPR \\
			\hline
			\cite{bassily2019private} & $\checkmark$ & $\checkmark$ & $\times$ & $\checkmark$ & $\times$ & $\mathcal{O}\left(\frac{\sqrt{m}}{n\epsilon}+\frac{1}{\sqrt{n}}\right)$ \\
			\hline
			\cite{feldman2020private} & $\checkmark$ & $\times$ & $\checkmark$ & $\checkmark$ & $\times$ & $\mathcal{O}\left(\frac{m}{n^2\epsilon^2}+\frac{1}{n}\right)$ \\
			\hline
			\cite{feldman2020private} & $\checkmark$ & $\times$ & $\times$ & $\checkmark$ & $\times$ & $\mathcal{O}\left(\frac{\sqrt{m}}{n\epsilon}+\frac{1}{\sqrt{n}}\right)$ \\
			\hline
			Ours & $\checkmark$ & $\checkmark$ & $\times$ & $\times$ & $\checkmark$ & $\mathcal{O}\left(\min\left\{\frac{m}{n^2\epsilon^2}+\frac{1}{n},\frac{\sqrt{m}}{n^{1.5}\epsilon}+\frac{m^{1.5}}{n^{2.5}\epsilon^3}\right\}\right)$ \\
			\hline
			Ours (PIDP-SGD) & $\checkmark$ & $\checkmark$ & $\times$ & $\times$ & $\checkmark$ &  $\mathcal{O}\left(\min\left\{\frac{km}{n^3\epsilon^2}+\frac{1}{n},\frac{km}{n^{2.5}\epsilon^2}+\frac{1}{n^{1.5}},\frac{(km)^{1.5}}{n^2\epsilon^3}+\frac{\epsilon}{\sqrt{km}n}\right\}\right)$ \\
			\hline
		\end{tabular}
		\begin{tablenotes}
			\item[1] $k$ is the number of data instances that affect the model significantly, mentioned above.
		\end{tablenotes}
	\end{threeparttable}
\end{table*}

\subsection{Utility Analysis}

\subsubsection{Excess Empirical Risk}

\begin{The}
	Suppose that $\ell(\theta,z)$ is $G$-Lipschitz, $L$-smooth, and satisfies PL condition over $\theta$.
	With learning rate $\alpha=\frac{1}{L}$ and $k$ data instances affect the model significantly, the excess empirical risk bound can be improved to:
	\begin{equation*}
		\mathbb{E}\left[L(\theta_{priv};D)-L^*\right] \leq
		\mathcal{O}\left(\frac{kmG^2\log(1/\delta_1)\log(n)}{n^3\epsilon_1^2}\right),
	\end{equation*}
	where the expectation is taken over the algorithm and dataset $D$, $T=\mathcal{O}(\log(n))$.
\end{The}

\begin{proof}
	As in the proof of Theorem 1, if there are $k$ data instances affect the model significantly, (\ref{ITERt}) can be written as:
	\begin{equation*}
		\begin{aligned}
			\mathbb{E}\left[L(\theta_{t+1};D)-L(\theta_t;D)\right]&\overset{(L)}{\leq}\mathbb{E}\left[\left<\nabla_\theta
			L(\theta_t;D),
			\theta_{t+1}-\theta_t\right>+\frac{L}{2}\|\theta_{t+1}-\theta_t\|_2^2\right] \\
			&\leq-\frac{1}{2L}\mathbb{E}_{D}\left[\|\nabla_\theta
			L(\theta_t;D)\|_2^2\right]+\frac{k}{n}\frac{1}{2L}\mathbb{E}\|b\|_2^2,
		\end{aligned}
	\end{equation*}
	where the last inequality holds because we have the probability $\frac{k}{n}$ to reach the data instance $z_t$ `meeting' the noise.
	
	Thus, we have:
	\begin{equation*}
		\begin{aligned}
			\mathbb{E}\left[L(\theta_{t+1};D)-L(\theta_t;D)\right]&\leq-\frac{\mu}{L}(\mathbb{E}_D\left[L(\theta_t;D)-L^*\right])+\frac{k}{n}\frac{m\sigma^2}{2L}.
		\end{aligned}
	\end{equation*}
	
	Then, summing over $T$ iterations, we have:
	\begin{equation*}
		\begin{aligned}
			\mathbb{E}[L(\theta_T;D)-L^*]
			&\leq(1-\frac{\mu}{L})^T(\mathbb{E}_D\left[L(\theta_0;D)-L^*\right])+\frac{km\sigma^2}{2Ln}\left(\sum_{i=0}^{T-1}(1-\frac{\mu}{L})^i\right) \\
			&\leq(1-\frac{\mu}{L})^T(\mathbb{E}_D\left[L(\theta_0;D)-L^*\right])+\frac{km\sigma^2}{2\mu n}.
		\end{aligned}
	\end{equation*}
	
	Taking
	$T=\mathcal{O}\left(\log\left(\frac{n^3\epsilon_1^2}{kmG^2}\right)\right)$ and
	$\sigma$ discussed before:
	\begin{equation*}
		\begin{aligned}
			\mathbb{E}[L(\theta_{priv};D)-L^*]&=\mathbb{E}[L(\theta_T;D)-L^*]\leq O\left(\frac{kmG^2\log(1/\delta_1)\log(n)}{n^3\epsilon_1^2}\right),
		\end{aligned}
	\end{equation*}
	where the expectation is taken over $D$ and the algorithm.
\end{proof}

Noting that $\frac{k}{n}<1$, we have:
\begin{equation*}
	\mathcal{O}\left(\frac{kmG^2\log(1/\delta)\log(n)}{n^3\epsilon^2}\right)<\mathcal{O}\left(\frac{mG^2\log(1/\delta)\log(n)}{n^2\epsilon^2}\right).
\end{equation*}
As a result, the excess empirical risk bound brought by Theorem 6 is better than Theorem 1.

\subsubsection{Excess Population Risk}

Like in Section \uppercase\expandafter{\romannumeral4}, we first analyze the generalization error.
Via the proof of Theorem 2, we find that the generalization error is only related to $\|\theta_i^T-\theta^T\|_2$ and $L(\theta_{priv};D)$, and these terms are only determined by the optimization process, so the generalization error of Algorithm 1 is the same as which given in Theorem 2.
Then we come to the excess population risk.

\begin{The}
	If $\ell(\cdot)$ is $G$-Lipschitz, $L$-smooth, and satisfies the PL condition over $\theta$.
	With learning rate $\alpha=\frac{1}{L}$, $T=\mathcal{O}(\log(n))$ and $k$ data instances affect the model significantly, the excess population risk can be improved to:
	\begin{equation*}
		\begin{aligned}
			&\mathbb{E}\left[L_\mathcal{P}\left(\theta_{priv}\right)-\min_{\theta}L_\mathcal{P}\left(\theta\right)\right] \\
			&\leq\mathcal{O}\left(\min\left\{\frac{km}{n^3\epsilon^2}+\frac{1}{n},\frac{km}{n^{2.5}\epsilon^2}+\frac{1}{n^{1.5}},\frac{(km)^{1.5}}{n^2\epsilon^3}+\frac{\epsilon}{\sqrt{km}n}\right\}\right).
		\end{aligned}
	\end{equation*}
\end{The}

The proof is similar to Theorem 3 and the discussion given in Remark 4.
The first, second, third and last terms on the right side of the inequality are derived from taking $\tau=\mathcal{O}(1),\mathcal{O}(\sqrt{n})$ and $\mathcal{O}(n\epsilon/\sqrt{km})$, respectively.

Noting that $\frac{k}{n}<1$, the result is better than which given by Theorem 3.


\begin{Rem}
	Note that $\epsilon$ and $\epsilon_1$ can be set almost the same, so we do not
	distinguish them in the discussion above.
\end{Rem}

For the first time, we theoretically prove that the excess risk of differential privacy models can be better by considering data heterogeneity.
It may give new inspirations to the utility analysis in the future work.

\section{Comparison with Related Work}

\subsection{Utility Bounds}

For excess empirical risk, by detailed theoretical analysis, we give a sharper bound for DP-SGD.
Our analyzed excess empirical risk bound of is better than which proposed by \cite{bassily2014private,wu2017bolt,wang2019DPLSSGD,bassily2019private} and achieves the best result $\mathcal{O}\big(\frac{m}{n^2\epsilon^2}\big)$.
For our proposed `performance improving' method: PIDP-SGD, our analyzed excess empirical risk bound is further tighter by a factor of $\mathcal{O}\big(\frac{k}{n}\big)$.
It is worth emphasizing that most of the methods proposed previously assume that the loss function is convex, which is not required in our method.
Under this circumstance, we achieve a better result, which is attractive.
Additionally, for the non-convex analysis given in \cite{wang2017differentially}, the excess empirical risk bound of our analyzed DP-SGD method is better by a factor of $\mathcal{O}\big(\log(n)\big)$ (as discussed in Remark 3) and the PIDP-SGD method is better by a factor of $\mathcal{O}\big(\frac{k\log(n)}{n}\big)$.
Details can be found in Table 1, in which $G,L$, S.C., C, PL represent $G$-Lipschitz, $L$-smooth, strongly convex, convex and PL inequality, respectively and EER, EPR denote the Excess Empirical Risk and the Empirical Population Risk, respectively.

Previous works always discuss the excess empirical risk but seldom analyze the excess population risk, here we compare our method with the best result $\mathcal{O}\big(\frac{m}{n^2\epsilon^2}+\frac{1}{n}\big)$ given in \cite{feldman2020private}, under strongly convex condition.
As shown in Table \uppercase\expandafter{\romannumeral1}, our result is better by a factor up to $\mathcal{O}\big(\frac{1}{\sqrt{n}}\big)$ under the best case.
When it comes to our proposed PIDP-SGD method, our result is further improved by a factor up to $\mathcal{O}\big(\frac{k}{n}\big)$.
Noting that the best result given in \cite{feldman2020private} requires the loss function to be strongly convex, which means that our result is not only better but also strictly more general than which given in \cite{feldman2020private}.
For the result under the convex assumption $\mathcal{O}\big(\frac{\sqrt{m}}{n\epsilon}+\frac{1}{\sqrt{n}}\big)$, our results (both the original one and the one given by PIDP-SGD) are much better.
Although it is hard to compare the convexity and the PL condition, our results can be applied to some of the non-convex models (as shown in Definition 5).

\begin{figure*}[ht]
	\vskip 0.2in
	\begin{center}
		\centering{
			\subfigure[KDDCup99 (LR)]{\includegraphics[width=0.3\textwidth]{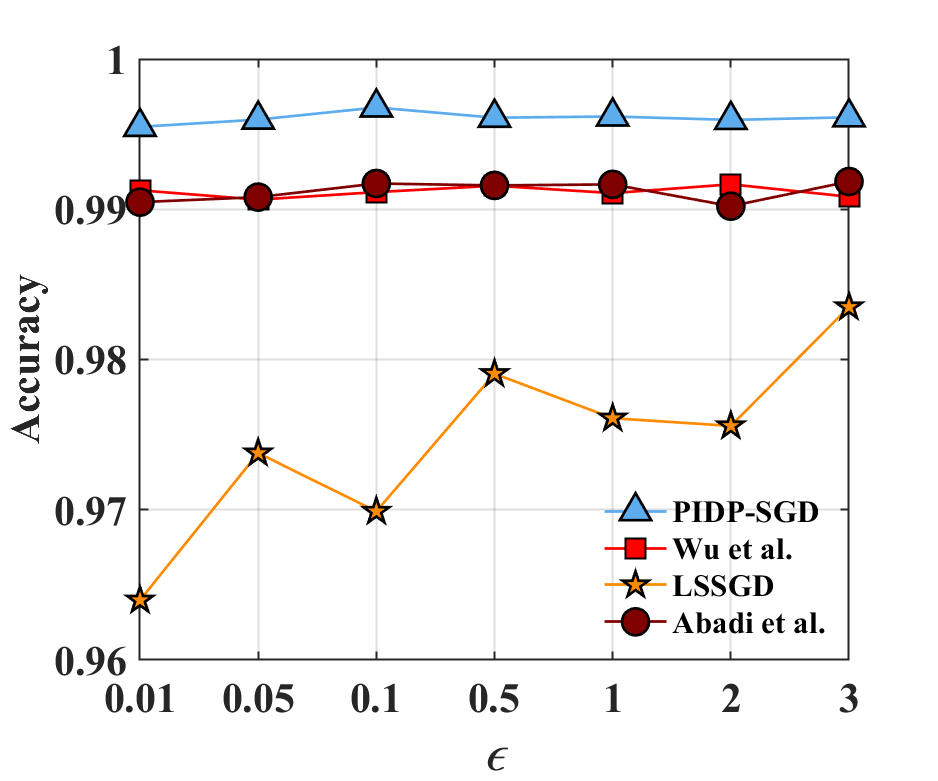}}
			\subfigure[Adult (LR)]{\includegraphics[width=0.3\textwidth]{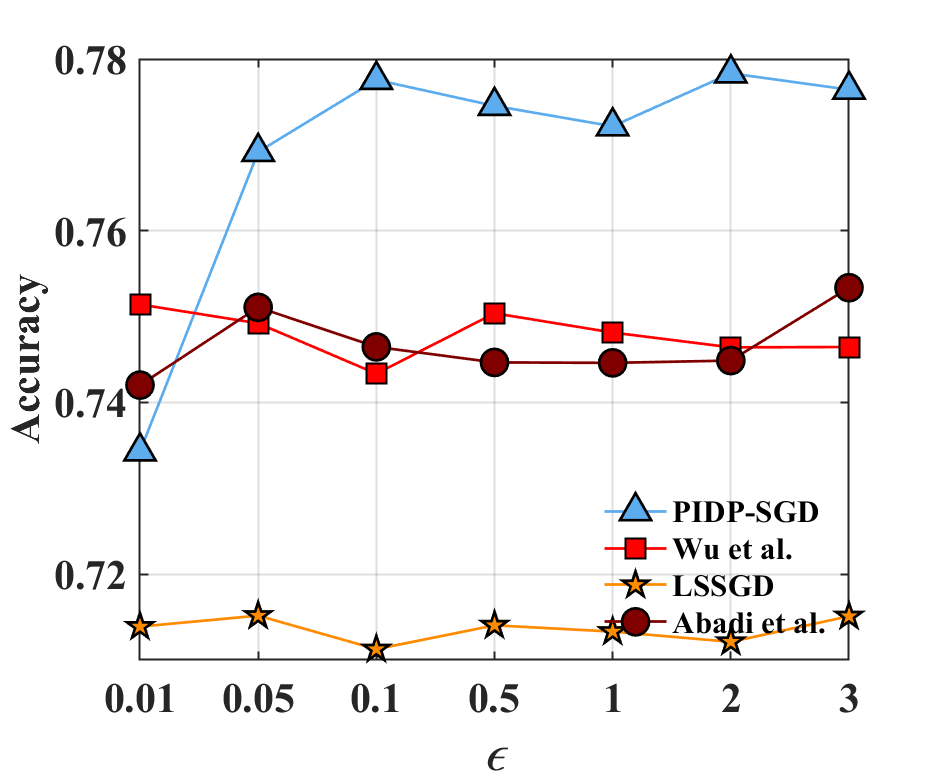}}
			\subfigure[Bank (LR)]{\includegraphics[width=0.3\textwidth]{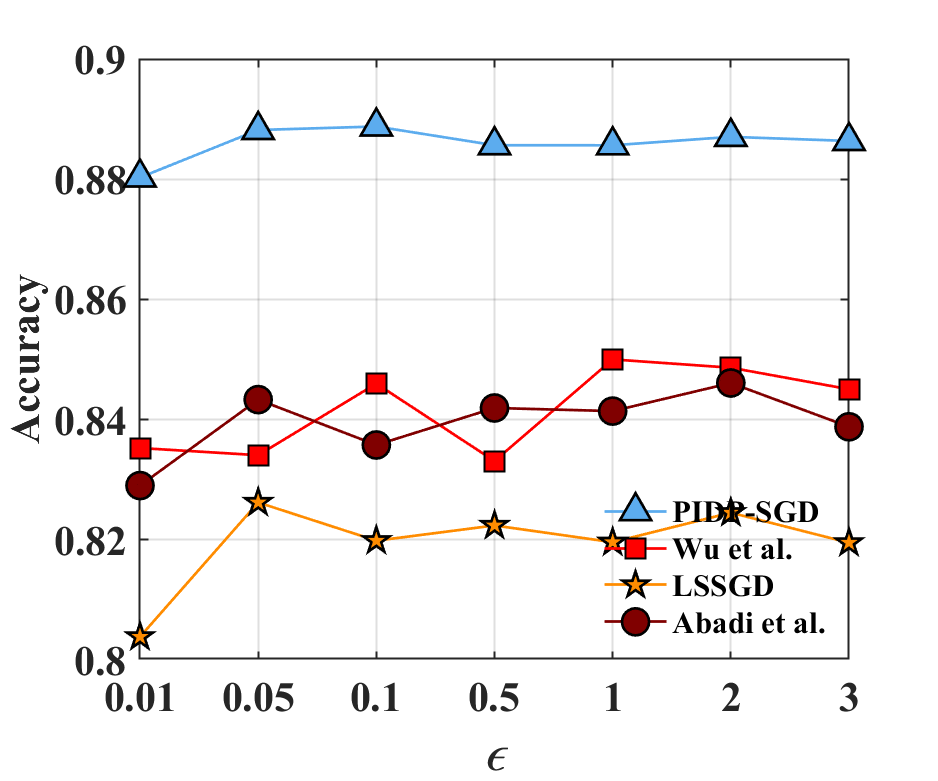}}
			\subfigure[KDDCup99
			(MLP)]{\includegraphics[width=0.3\textwidth]{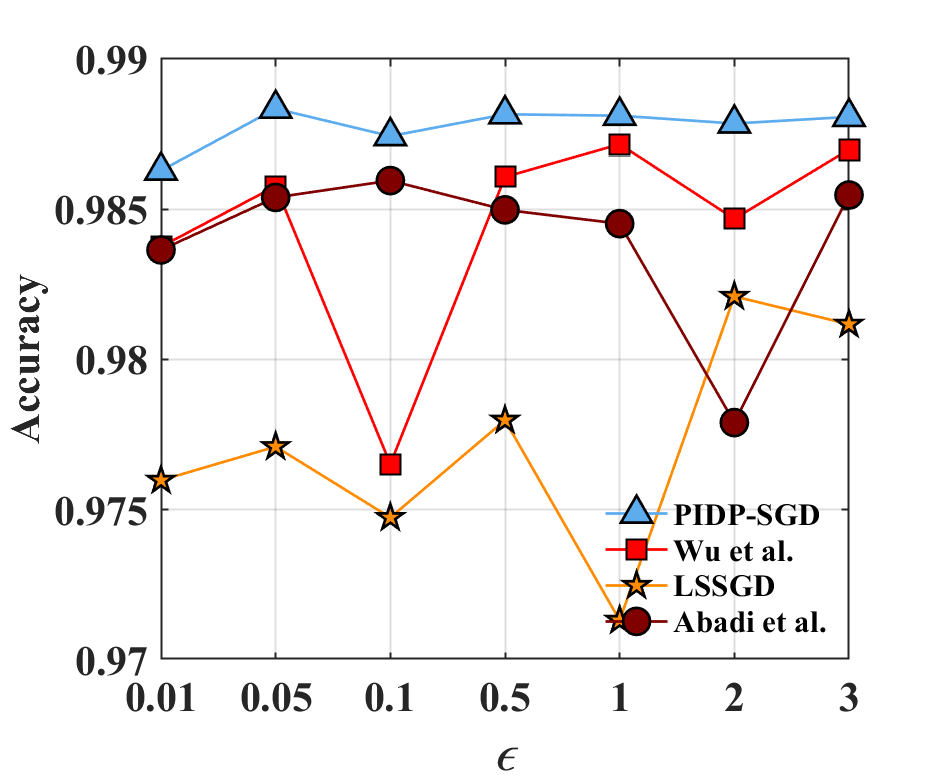}}
			\subfigure[Adult (MLP)]{\includegraphics[width=0.3\textwidth]{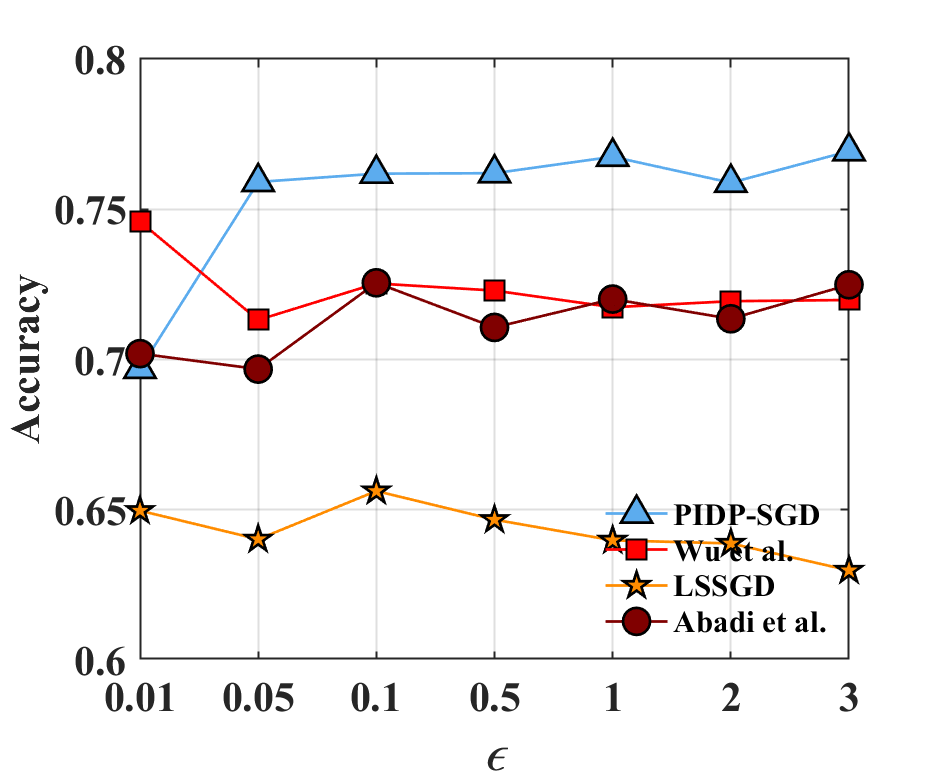}}
			\subfigure[Bank (MLP)]{\includegraphics[width=0.3\textwidth]{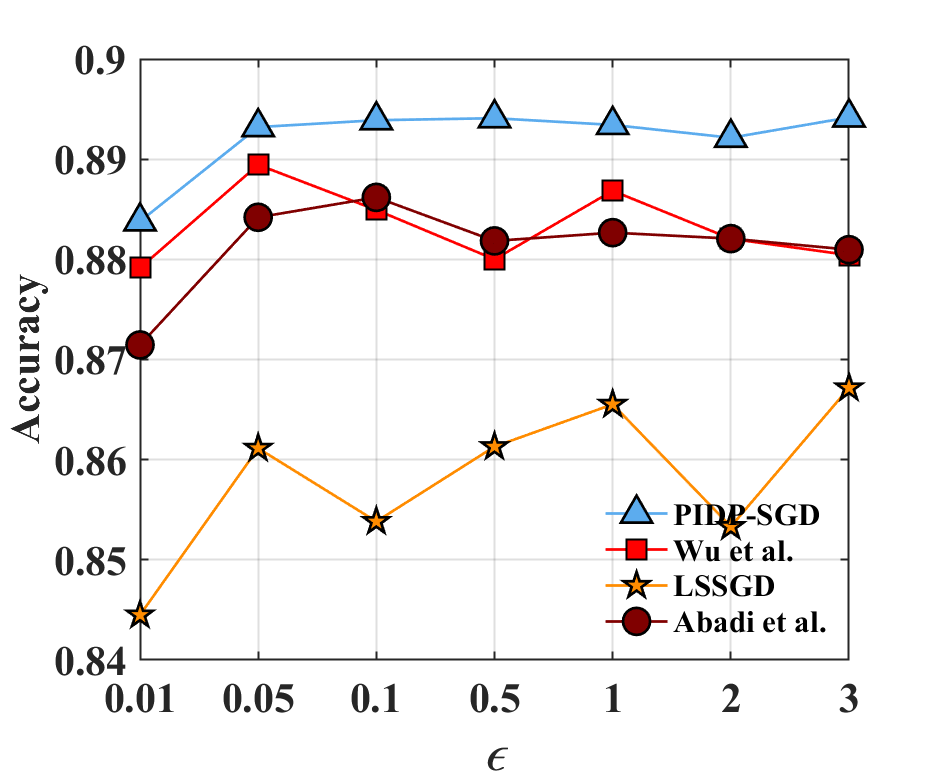}}
		}
		\caption{Accuracy over $\epsilon$, LR denotes logistic regression model and MLP denotes the deep learning model.}
		\label{Accuracy}
	\end{center}
	\vskip -0.2in
\end{figure*}

\begin{figure*}[ht]
	\vskip 0.2in
	\begin{center}
		\centering{
			\subfigure[KDDCup99
			(LR)]{\includegraphics[width=0.3\textwidth]{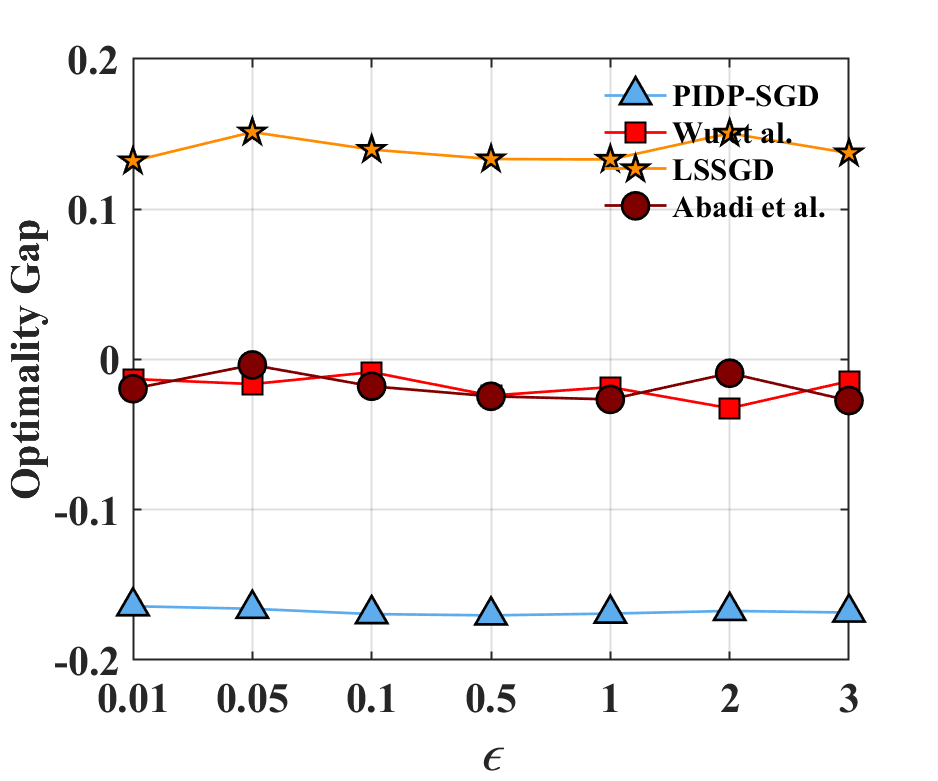}}
			\subfigure[Adult
			(LR)]{\includegraphics[width=0.3\textwidth]{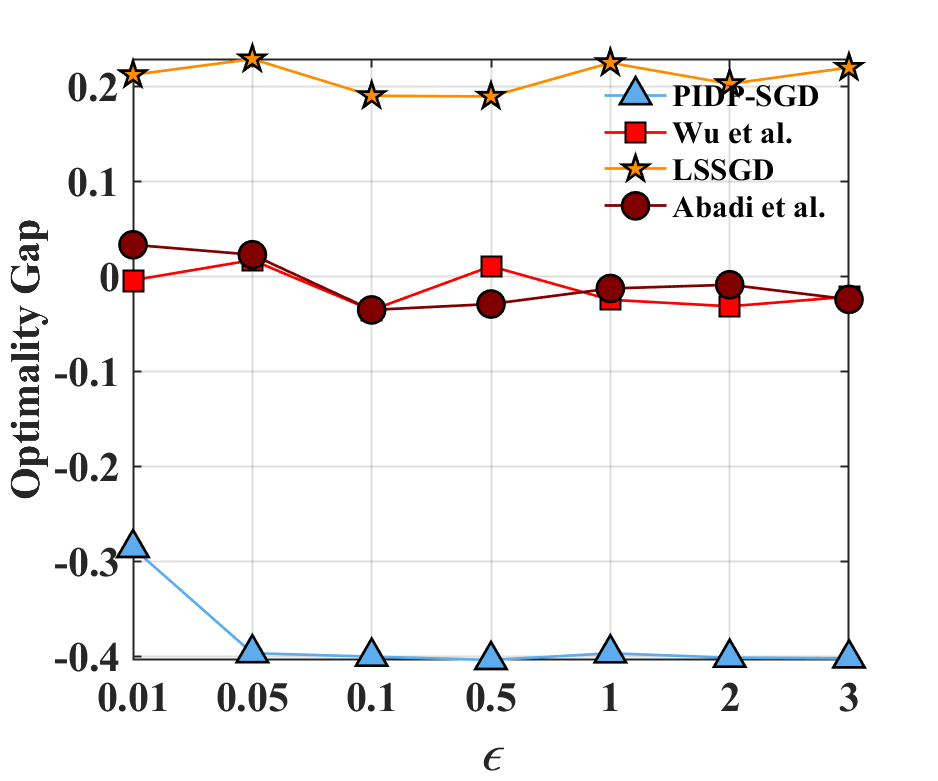}}
			\subfigure[Bank (LR)]{\includegraphics[width=0.3\textwidth]{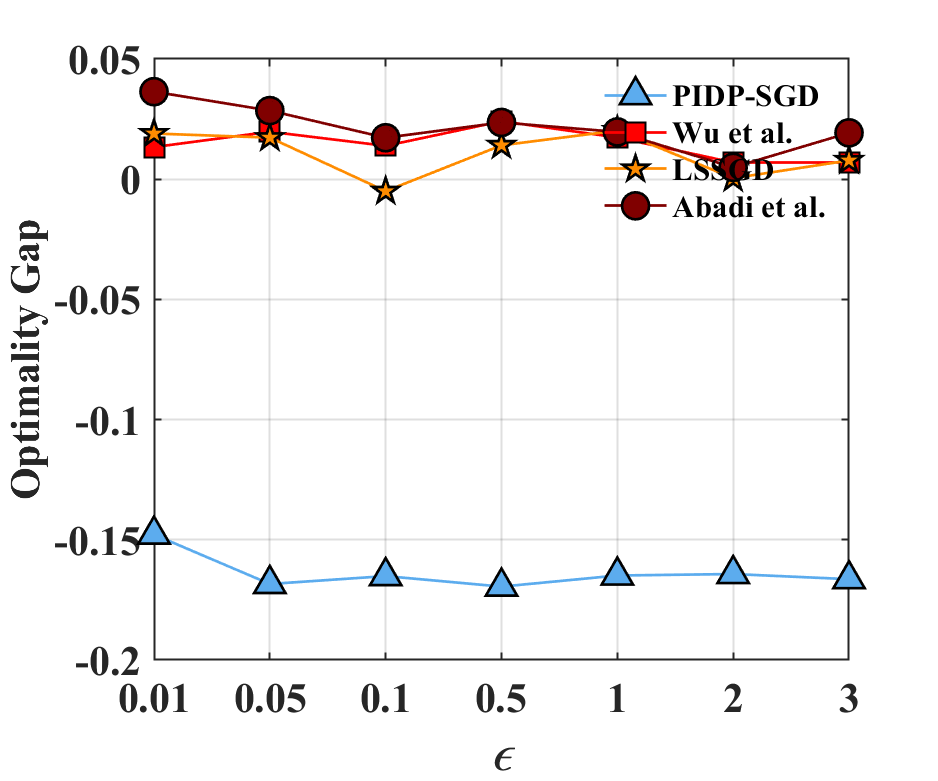}}
			\subfigure[KDDCup99
			(MLP)]{\includegraphics[width=0.3\textwidth]{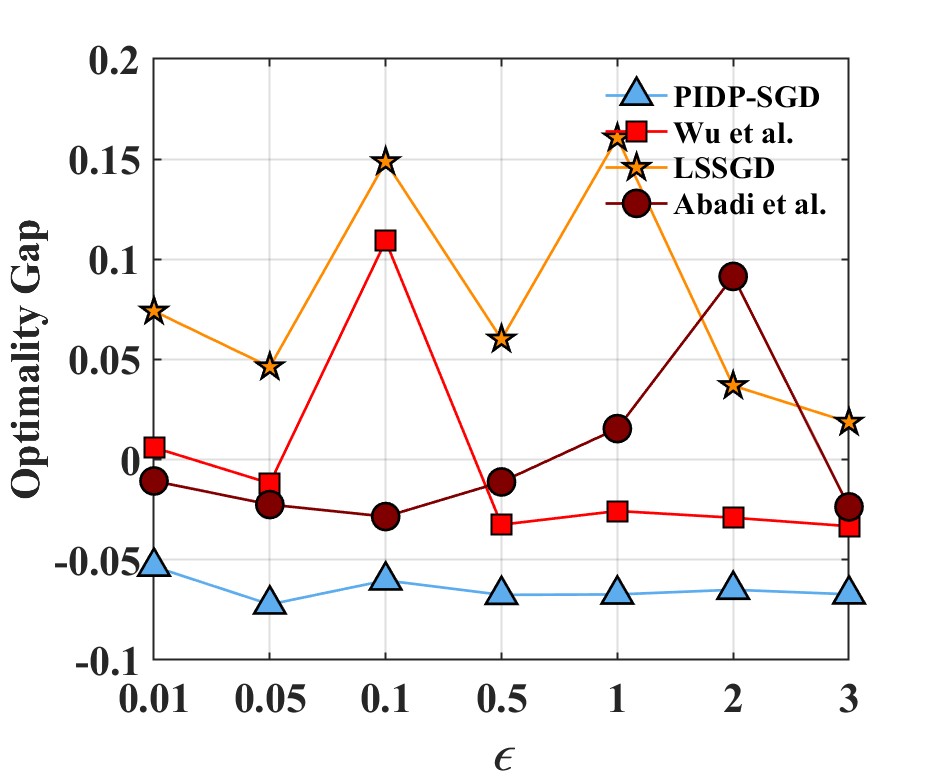}}
			\subfigure[Adult
			(MLP)]{\includegraphics[width=0.3\textwidth]{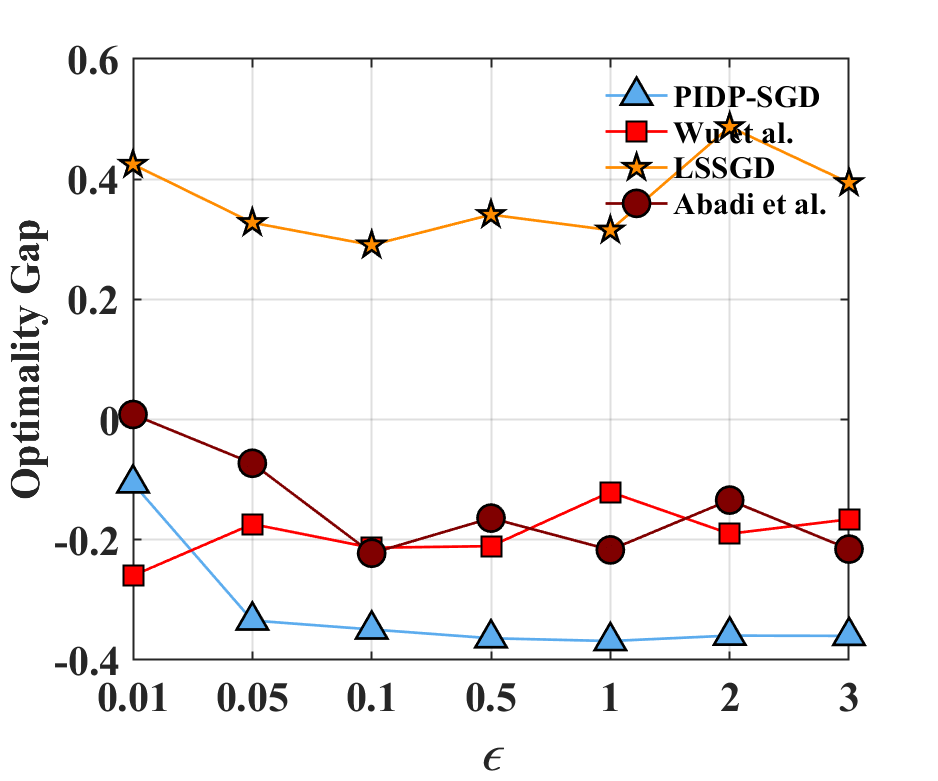}}
			\subfigure[Bank
			(MLP)]{\includegraphics[width=0.3\textwidth]{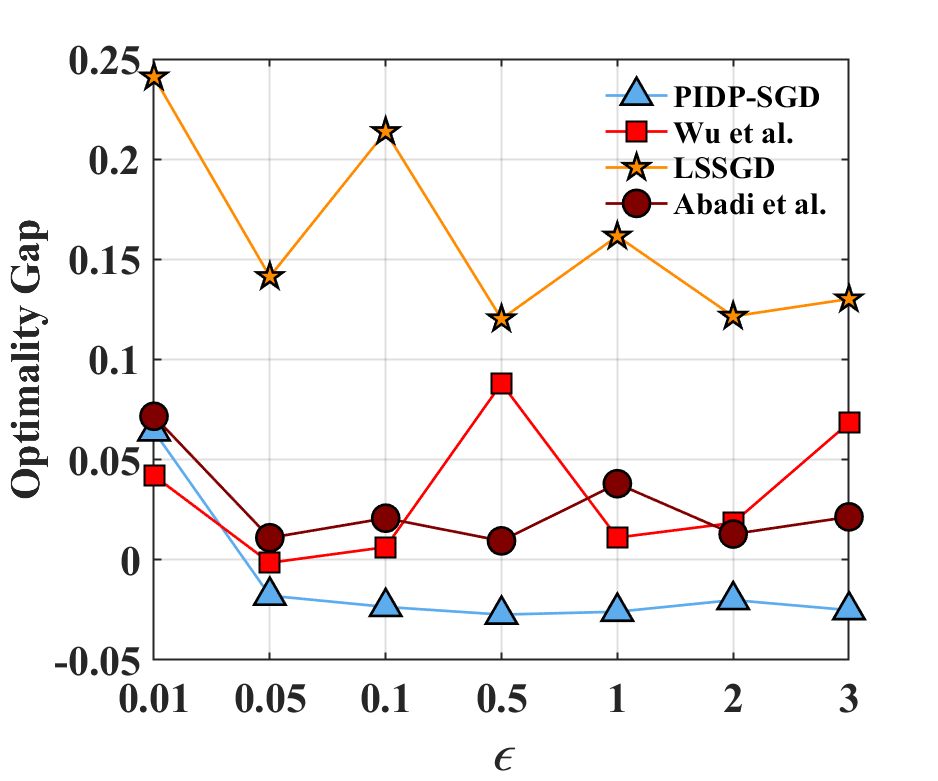}}
		}
		\caption{Optimality gap over $\epsilon$, LR denotes logistic regression model and MLP denotes the deep learning model.}
		\label{Optimality gap}
	\end{center}
	\vskip -0.2in
\end{figure*}

\section{Experimental Results}

Experiments on several real datasets are performed on the classification task.
Since our method is based on SGD, we compare our method with previous DP-SGD methods.
Specifically, we compare our method with the gradient perturbation method proposed in \cite{abadi2016deep}, the output perturbation method proposed in \cite{wu2017bolt} and the DP-LSSGD method proposed in \cite{wang2019dp}.
The performance is measured in terms of classification accuracy and the optimality gap.
The accuracy represents the performance on the testing set, and the optimality gap represents the excess empirical risk on the training set.
The optimality gap is denoted by $L(\theta_{priv};D)-L^*$.

We use both logistic regression model and deep learning model on the datasets KDDCup99 \cite{KDD}, Adult \cite{Adult&Iris} and Bank \cite{Bank}, where the total number of data instances are 70000, 45222, and 41188, respectively.
In the experiments, to make the model satisfies the assumptions (such as PL condition) mentioned in the theoretical part, the deep learning model is denoted by Multi-layer Perceptron (MLP) with one hidden layer whose size is the same as the input layer.
Training and testing datasets are chosen randomly.

In all the experiments, total iteration rounds $T$ is chosen by cross-validation.
For our performance improving method, we set $RT_{local}=T$.
We evaluate the performance of our proposed PIDP-SGD method and some of previous algorithms over the differential privacy budget $\epsilon$.
For $\epsilon$, we set it from 0.01 to 3, and in the PIDP-SGD method, we set $\epsilon_1=3\epsilon_2=3\epsilon/4$ to guarantee $\epsilon_1+\epsilon_2=\epsilon$.
The results are shown in Figure~\ref{Accuracy} and Figure~\ref{Optimality gap}.

Figure~\ref{Accuracy} shows that as the privacy budget $\epsilon$ increases, so does the accuracy, which follows the intuition.
When applying the PIDP-SGD algorithm, the accuracy rises on most datasets, which means that our proposed `performance improving' method is effective.
Meanwhile, when $\epsilon$ is small, the difference (on accuracy) between traditional methods and `performance improving' method is also small.
However, as $\epsilon$ increases, the `performance improving' method becomes more and more competitive.
The reason is that larger $\epsilon$ means that more data instances `escape' the injected noise, leading to better accuracies.

Figure~\ref{Optimality gap} shows that on some datasets, by applying PIDP-SGD algorithm, the optimality gap of our method is almost 0, which means that it achieves almost the same performance as the model without privacy in some scenarios.
Besides, similar to the accuracy in Figure~\ref{Accuracy}, the optimality gap decreases as $\epsilon$ increases, which follows our intuition.
Moreover, on some datasets, the performance of some of the methods fluctuates.
The reason is that in the setting of differential privacy, random noise is injected into the model, so it is a common phenomenon.

Additionally, on some datasets, the performance of our `performance improving' method is worse when $\epsilon$ is small, the reason is that part of the privacy budget is allocated to $c_t$ (in our experiments, we set $\epsilon_1=3\epsilon_2=3\epsilon/4$, so the ratio is $1/4$), which means `pure privacy budget' on the model is smaller.
Thus, with the increase of $\epsilon$, the `performance improving' method becomes more competitive, which has been analyzed before in Section \uppercase\expandafter{\romannumeral5}.$C$.
Experimental results show that our proposed PIDP-SGD algorithm significantly improves the performance under most circumstances.

\section{Conclusions}
In this paper, we give sharper excess risk bounds of traditional DP-SGD paradigm, including the excess empirical risk bound and the excess population risk bound.
Theoretical results show that our given excess risk bounds achieve better excess risk bounds than some of the best previous methods under non-convex condition, and even better than previous results under convex condition.
Meanwhile, based on DP-SGD, we attempt to improve the performance from a new perspective: considering data heterogeneity, rather than treating all data the same.
In particular, we introduce the influence function (IF) to DP-SGD to analyze the contribution of each data instance to the final model, and the approximation error analysis shows that IF is reasonable to approximate the contribution.
In this way, we propose the PIDP-SGD algorithm: only adding noise to the data demonstrating significant contributions (more than $e^\epsilon$) when training.
Detailed theoretical analysis and experimental results show that our proposed `Performance Improving' algorithm achieves better performance, without the convex assumption.
Moreover, the new perspective from the view of treating different data instances differently may give new inspirations to future work, including the privacy analysis and the utility analysis.


%
%
%
\bibliographystyle{splncs04}
\bibliography{iccs}
%




\end{document}